\documentclass[10pt,twocolumn,letterpaper]{article}
\usepackage[table,xcdraw,dvipsnames]{xcolor}
\usepackage[pagenumbers]{iccv} 
\usepackage{multirow}
\usepackage{stfloats}

\usepackage{amsmath}
\usepackage{bbm}
\usepackage{makecell}
\usepackage{marvosym}
\definecolor{iccvblue}{rgb}{0.21,0.49,0.74}
\usepackage[pagebackref,breaklinks,colorlinks,allcolors=iccvblue]{hyperref}

\title{SDD-4DGS: Static-Dynamic Aware Decoupling in Gaussian Splatting for 4D Scene Reconstruction}

\author{Dai Sun$^\&$, Huhao Guan$^\&$, Kun Zhang$^\&$, Xike Xie$^\textsuperscript{\Letter}$, S. Kevin Zhou$^\textsuperscript{\Letter}$
\thanks{$^\&$: Authors contribute equally to this work, \textsuperscript{\Letter}: Corresponding authors}\\
School of Biomedical Engineering, Division of Life Sciences and Medicine, \\University of Science and Technology of China, Hefei, Anhui, 230026, P.R.China\\
}

\begin{document}
\maketitle
\begin{abstract}
Dynamic and static components in scenes often exhibit distinct properties, yet most 4D reconstruction methods treat them indiscriminately, leading to suboptimal performance in both cases. This work introduces SDD-4DGS, the first framework for static-dynamic decoupled 4D scene reconstruction based on Gaussian Splatting.
Our approach is built upon a novel probabilistic dynamic perception coefficient that is naturally integrated into the Gaussian reconstruction pipeline, enabling adaptive separation of static and dynamic components. With carefully designed implementation strategies to realize this theoretical framework, our method effectively facilitates explicit learning of motion patterns for dynamic elements while maintaining geometric stability for static structures.
Extensive experiments on five benchmark datasets demonstrate that SDD-4DGS consistently outperforms state-of-the-art methods in reconstruction fidelity, with enhanced detail restoration for static structures and precise modeling of dynamic motions. The code will be released.
\end{abstract}
\vspace{-20pt}    
\section{Introduction}
\label{sec:intro}

Reconstruction technology is vital in our daily lives, with spread used in films, VR applications, and driver assistance systems, highlighting its importance in computer vision and graphics~\cite{tewari2022advances, lu2024image}.
Reconstruction aims to estimate a 3D scene representation from multiple 2D perspectives, facilitating realistic rendering from novel views.
In contrast to conventional 3D reconstruction, which merely focuses on static scenes, 4D reconstruction constructs both the scene and its temporal information from scarce 2D observations, enabling effective modeling of the dynamic world commonly found in everyday life, thereby offering enhanced generalizability~\cite{yunus2024recent, ingale2021real}. 
Tech challenges arise since 4D reconstruction requires an additional complexity of temporal sequence modeling beyond 3D.

Recently, many impactful 4D reconstruction methods~\cite{duan20244d, wu20244d, li2024spacetime, yang2023real} try incorporating temporal information into the 3D Gaussian Splatting (3DGS)~\cite{kerbl20233d}, a high-fidelity rendering technique with real-time capabilities, to model dynamic scenes, showing highly promising performance. Typically, they focus on devising sophisticated modules for estimating dynamic information, \eg, Deformation Net~\cite{wu2022d}, and Polynomial-based Fitting~\cite{li2024spacetime}. Despite these advances, current research generally assumes that all components in a scene experience motion over time, overlooking the need to determine \textit{which specific components actually require dynamic modeling}.

\begin{figure*}[htbp]
    \centerline{\includegraphics[width=.95\linewidth]{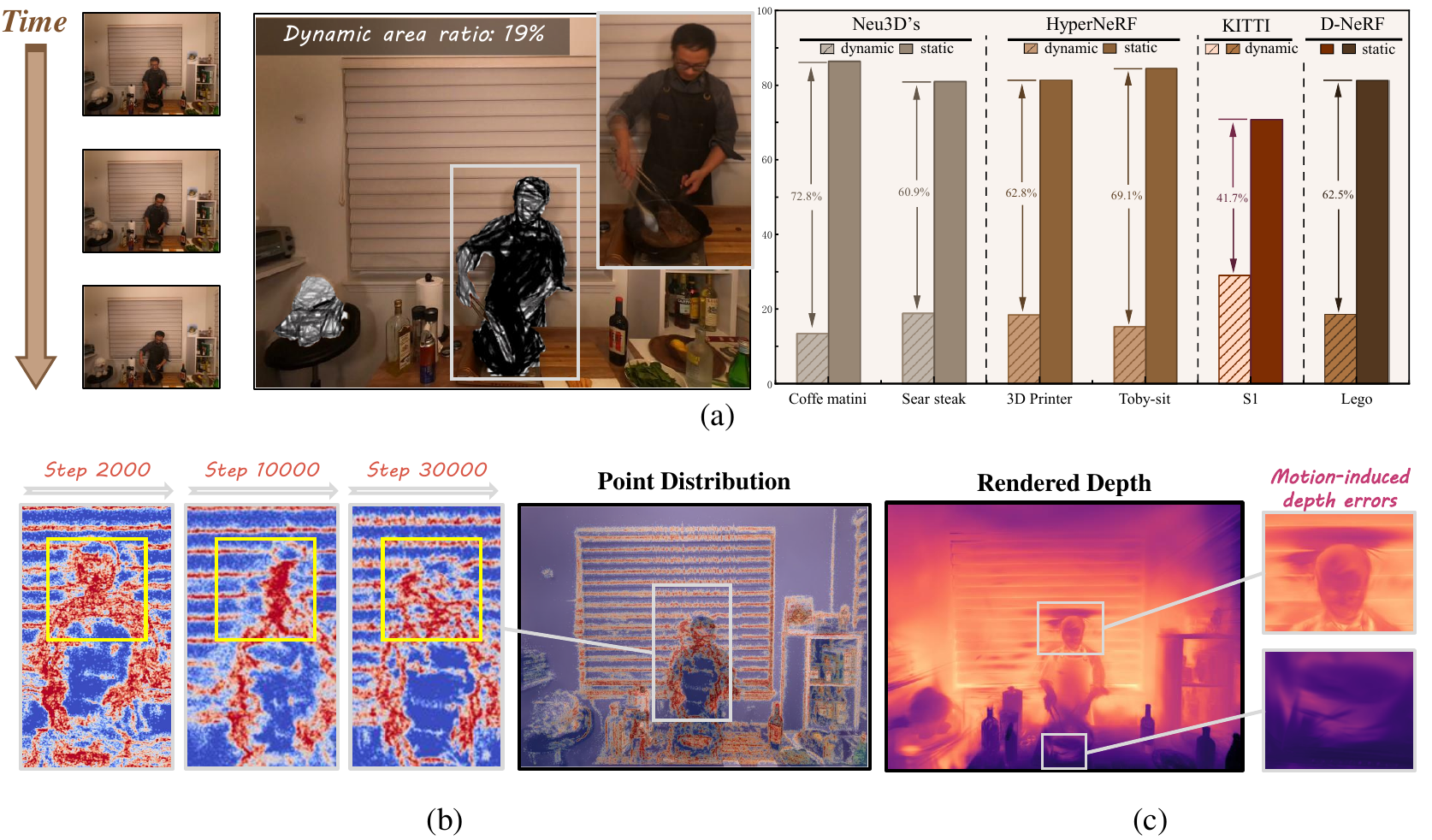}}
    \vspace{-7px}
    \caption{
    \textbf{Challenges in 4D scene reconstruction:} (a) Scenes typically consist of both static (e.g., buildings, walls) and dynamic (e.g., moving vehicles, pedestrians) components.  (b) The red pixels represent the distribution of points projected onto the imaging plane by Gaussians. The darker the color, the more Gaussians are projected onto the pixel. As the iterative optimization proceeds, the points describing the dynamic information of the head are gradually attracted to the surrounding static area. (c) The presence of dynamic objects introduces occlusions and uneven lighting on static components, leading to artifacts and reducing the quality of static scene reconstruction. Static background depth estimation errors due to head movement. \textit{Note:} Both (b) and (c) are derived from 4DGS~\cite{wu20244d}.
    }
    \vspace{-7px}  
    \label{fig:intro}
\end{figure*}

In reality, scenes are not solely composed of dynamic components; rather, static components such as buildings, streets, and walls often constitute a significant portion of the scene~\cite{cordts2016cityscapes}.
As depicted in Fig.~\ref{fig:intro}(a), our analysis reveals that in typical scenes, static structures occupy approximately $75-85\%$ of the scene volume, with dynamic objects (e.g., pedestrians, vehicles) confined to a much smaller spatial region, typically less than $20\%$ of the visible area.
Based on this crucial observation, we argue that existing methods ignore the inherent difference between dynamic and static components within a scene. Using a static/dynamic agnostic modeling approach for reconstruction inevitably leads to mutual limitations between dynamic and static information, whose disadvantages are two-fold.


\textbf{Firstly}, \textit{static low-frequency information impedes the optimization of dynamic high-frequency information, leading to less effective dynamic modeling}.
In 4D reconstruction models, static regions often exhibit consistent characteristics over time (such as fixed textures and geometric boundaries), remaining relatively stable in modeling. Therefore, static regions are likely to be low-loss and easily optimized during the optimization process~\cite{zhang2021understanding}.
Since reconstruction models are optimized by minimizing the loss function, the model iteratively tends to fit static regions, driving parameters closer to these stable low-loss areas.
As a result, static regions act as "low-energy anchors" in the model’s optimization, guiding the reconstruction of dynamic objects toward these regions. This causes the optimization of dynamic parameters to be heavily influenced by static information, thereby paying less attention to dynamic objects. 
For example, as shown in Fig.~\ref{fig:intro}(b), we find that during the iterative optimization progresses, the points describing the dynamic information of the head (marked in the yellow box) are gradually attracted to the surrounding static areas, becoming increasingly blurred and difficult to model its dynamic information.



\textbf{Secondly}, \textit{in dynamic environments, moving objects, by their nature, occlude static regions, which results in uneven lighting scenarios.} Specifically, 
the movement of dynamic objects can alter the direction and intensity of illumination, thus impacting the color appearance of static objects. 
Moreover, the high-speed movement of dynamic objects is prone to generating artifacts or motion blur during interactions with static objects. 
As a result, dynamic objects detrimentally influence the detail and geometric analysis of static objects, as shown in Fig.~\ref{fig:intro}(c) that the estimated depth of the static background behind the dynamic head is inaccurate.


We propose SDD-4DGS, a 4D Gaussian splatting framework with Static-Dynamic aware Decoupling capabilities to address the issues above. To our knowledge, we are the first to reconstruct 4D scenes from the perspective of static-dynamic aware decoupling using Gaussian Splatting. Our approach integrates two complementary innovations: (1) a theoretically rigorous dynamic perception coefficient derived in probabilistic form that naturally integrates into the Gaussian reconstruction pipeline, providing the mathematical foundation for component separation; and (2) a set of novel optimization strategies that effectively realize this theoretical potential in practical scenarios, addressing key implementation challenges that emerge when applying probabilistic models to complex dynamic scenes. In this manner, scenes are categorized as dynamic or static through a principled probabilistic framework and reconstructed with carefully designed optimization strategies. The integration of theory and practice makes SDD-4DGS a comprehensive solution for 4D reconstruction, greatly enhancing both accuracy and detail restoration across various reconstruction scenarios.

The contributions are summarized as follows:
\begin{itemize}
    \item To our best knowledge, SDD-4DGS is the first framework to achieve static-dynamic aware decoupling in 4D reconstruction using Gaussian Splatting~\cite{kerbl20233d}. 

    \item We propose a novel dynamic perception coefficient rigorously derived in a probabilistic form.  
    
    \item We present innovative strategies to tackle implementation optimization challenges while ensuring theoretical integrity, featuring a progressive constraint schedule and an automatic supervision mechanism.

    \item We conduct extensive evaluations on five datasets~\cite{pumarola2021d, park2021hypernerf, park2021nerfies, li2022neural,menze2015object} under four distinct experimental settings, achieving comparable or superior performance than previous state-of-the-art (SOTA) methods.

\end{itemize}






\section{Related Works}
\label{sec:related}

\noindent\textbf{Dynamic scene reconstruction} aims to recover the dynamic scene using time-sequenced images. Traditional approaches employ point clouds~\cite{liu2019meteornet, huang2022dynamic}, meshes~\cite{dong2018efficient,kim2012outdoor}, voxels~\cite{zhang2017mixedfusion,liu2020neural}, and light fields~\cite{gotardo2015photogeometric,kaveti2020removing}. 
Contemporary methods leverage machine learning, notably through NeRF-based and Gaussian splatting-based techniques. NeRF-based methods~\cite{cao2023hexplane,ost2021neural,alfonso2024dynerfactor} deliver photorealistic rendering via learnable differential volume rendering~\cite{mildenhall2021nerf}, yet face challenges with non-rigid deformations~\cite{xie2023deform2nerf}, occlusions~\cite{zhu2023occlusion}, and time-varying lighting~\cite{rudnev2022nerf}.
Gaussian splatting~\cite{kerbl20233d} represents 3D scenes through spatial distributions of Gaussian ellipsoids, offering quality and speed advantages. For dynamic scenes, approaches such as 4DGS~\cite{wu20244d} and Spacetime~\cite{li2024spacetime} incorporate temporal data, treating time as a dimension or employing polynomial bases for motion. Solutions such as Realtime~\cite{yang2023real} and 4DRotor~\cite{duan20244d} utilize time-integrated 4D covariance matrices. However, these often neglect the stable static background, which is crucial for preserving fixed global geometric integrity.

\noindent\textbf{Static-dynamic aware decoupling} methods decompose the scene into static and dynamic regions. 
Mask-based approaches~\cite{gao2021dynamic,yan2023nerf,yang2023ppr,liu2023robust} utilize binary masks to decouple dynamic and static components in scenes, improving reconstruction quality and handling specular effects distinctly.
Recent self-supervised methods~\cite{li2021neural,chen2022flow,wu2022d,tschernezki2021neuraldiff} achieve dynamic-static scene decomposition without reliance on motion masks, enhancing reconstruction fidelity through various novel loss functions and learning mechanisms.
In addition, some research has focused on synthesizing static components from dynamic scenes. The pioneering work NeRF On-the-go~\cite{ren2024nerf} handles complex occlusions in scenes by applying DINOv2~\cite{oquab2023dinov2} features to predict uncertainty, although it requires extended training times. The latest work, WildGaussians~\cite{kulhanek2024wildgaussians}, significantly improves the optimization speed based on 3DGS and analyzes lighting variations. However, these methods do not simultaneously model dynamic and static components and are thus not directly applicable for dynamic scene reconstruction. To the best of our knowledge, our SDD-4DGS is the first to introduce a self-supervised dynamic-static decoupling mechanism within a Gaussian-Splatting-based dynamic reconstruction approach, effectively extending 4D reconstruction methods.

\begin{figure*}[htbp]
    \centerline{\includegraphics[width=\linewidth]{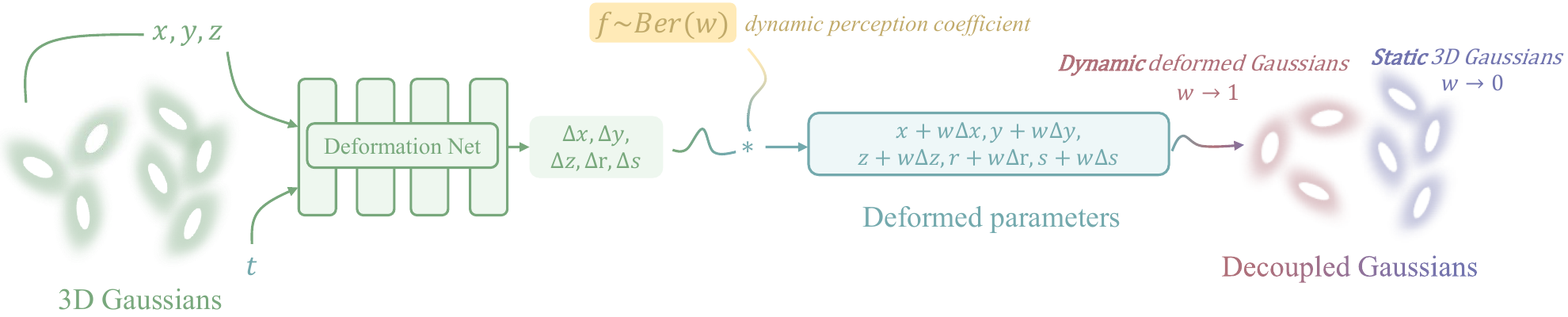}}
    \vspace{-5px}  
    \caption{\textbf{Overview of the proposed SDD-4DGS pipeline for static-dynamic decoupling in 4D reconstruction.}  The framework decouples static and dynamic components by integrating a novel dynamic perception coefficient into 4DGS~\cite{wu20244d}. The pipeline involves several key stages: initialization of 3D Gaussians, computation of deformation parameters through a deformation network, and dynamic regulation using the dynamic perception coefficient. Gaussians are then decoupled into static and dynamic groups, each optimized separately through loss functions which are detailed in Sec.~\ref{subsubsec:Uncertainty Guided Decoupling}. }
    \vspace{-12px}  
    \label{fig:pipeline}
\end{figure*}

\section{Methodology}\label{sec:methodology}
In this section, we begin by briefly reviewing 3D Gaussian Splatting (3DGS)~\cite{kerbl20233d} and its extension for dynamic scenes, 4D Gaussian Splatting (4DGS)~\cite{wu20244d}, in Sec.~\ref{subsec:Preliminary}. Then, in Sec.~\ref{subsec:Separation}, we introduce SDD-4DGS for 4D scene reconstruction. Finally, in Sec.~\ref{subsec:Optimization}, we introduce our detailed optimization objectives.

\subsection{Preliminary of 3D $\And$ 4D Gaussian Splatting}\label{subsec:Preliminary}

The 3D and 4D Gaussian Splatting methods efficiently reconstruct static and dynamic scenes. The 3D approach accurately represents static structures, while the 4D approach captures dynamic changes by incorporating temporal dimensions, facilitating advanced spatiotemporal modeling.


\noindent\textbf{3D Gaussian Splatting} (3DGS)~\cite{kerbl20233d} represents a 3D scene as a set of Gaussian ellipsoids, each $\boldsymbol{x} \in \mathbb{R}^3 \sim \mathcal{N}(\boldsymbol{\mu}, \boldsymbol{\Sigma})$ defined by a mean $\boldsymbol{\mu} \in \mathbb{R}^3$ and covariance $\boldsymbol{\Sigma}=\boldsymbol{R}\boldsymbol{S}\boldsymbol{S}^T\boldsymbol{R}^T$. Here, $\boldsymbol{S}=\operatorname{diag}(s_x, s_y, s_z)$ and $\boldsymbol{R}$ are quaternion-based. Furthermore, the color $c$ using spherical harmonics (SH) and an opacity parameter $\alpha$ is incorporated.
In sum, each 3D Gaussian ellipsoid is represented as $\mathcal{G}=$ $\left\{\boldsymbol{\mu}, \boldsymbol{S}, \boldsymbol{R}, c, \alpha\right\}$.
For rendering, 3DGS projects $\boldsymbol{x}$ to 2D $\boldsymbol{x}^{2d} \sim \mathcal{N}(\boldsymbol{\mu}^{2d}, \boldsymbol{\Sigma}^{2d})$ via approximate projection.
And then, integrating color $c$ and opacity $\alpha$, each pixel is shaded through \(\alpha\)-blending:
\vspace{-5px}
\begin{equation}\label{eq:iuv}
\resizebox{.9\hsize}{!}{
$
\begin{aligned}
    \mathcal{I}(u, v)= 
    \sum_{i=1}^N p_i\left(u,v\right) \alpha_i c_i\left(d_i\right) \prod_{j=1}^{i-1}\left(1-p_j\left(u,v\right) \alpha_j\right),
\end{aligned}
$
}
\vspace{-5px}
\end{equation}
where \( I(u, v) \) represents the pixel color at the position \( (u, v) \) within the image plane; $p_i\left(u,v\right)$ denotes the 2D projection $\boldsymbol{x}_i^{2d} \sim \mathcal{N}(\boldsymbol{\mu}_i^{2d}, \boldsymbol{\Sigma}_i^{2d})$ of the  \( i \)-th 3D Gaussian ellipsoid $\boldsymbol{x}_i \sim \mathcal{N}(\boldsymbol{\mu}_i, \boldsymbol{\Sigma}_i)$, indicating the probability density at pixel \( (u, v) \), defined as \( p_i\left(u,v\right) = p \left(u, v; \boldsymbol{\mu}^{2d}_i, \boldsymbol{\Sigma}^{2d}_i\right) \); \( c_i(d_i) \) indicates the color of the \( i \)-th visible Gaussian ellipsoid observed from the viewing direction \( d_i \).

\noindent\textbf{4D Gaussian Splatting} (4DGS)~\cite{wu20244d} extends the Eq.~(\ref{eq:iuv}) by introducing a timestamp \( t \), thereby linking the color of each pixel not only with the spatial information of the projected 3D Gaussian $\boldsymbol{x}$ but also with the temporal information. As a result, the pixel color representation in 3DGS, expressed as $\mathcal{I}(u, v)$, is extended to $\mathcal{I}(u, v, t)$, leading to:
\vspace{-5px}  
\begin{equation}\label{eq:I(uvt)}
\resizebox{.9\hsize}{!}{
$
    \mathcal{I}(u, v, t)=\sum_{i=1}^N p_i(u, v, t) \alpha_i c_i(d) \prod_{j=1}^{i-1}\left(1-p_j(u, v, t) \alpha_j\right),    
$
}
\vspace{-5px}
\end{equation}
where \( p_i(u, v, t) \) denotes the probability density  of the  \( i \)-th projected 3D Gaussian ellipsoid at pixel \( (u, v) \) at the current time \( t \). To model the Gaussian distribution that changes over time, 4DGS~\cite{wu20244d} introduces Gaussian parameter corrections based on temporal dependencies, defined as:
\vspace{-5px}  
\begin{equation}\label{eq:4dgs Delta}
    \boldsymbol{\mu}_{t} = \boldsymbol{\mu}_0 + \Delta \boldsymbol{\mu}_t, 
    \boldsymbol{\Sigma}_{t} = \boldsymbol{\Sigma}_0 + \Delta \boldsymbol{\Sigma}_t,
    \vspace{-5px}  
\end{equation}
where, \( \boldsymbol{\mu}_0 \) and \( \boldsymbol{\Sigma}_0 \) represent the original 3D Gaussian distribution within the canonical space, while \( \Delta \boldsymbol{\mu}_t \) and \( \Delta \boldsymbol{\Sigma}_t \) are dynamically adjusted according to the timestamp \( t \) using a spatio-temporal structure encoder and a multi-head Gaussian deformation decoder. Upon obtaining the deformed 3D Gaussian $\boldsymbol{x}_t\sim \mathcal{N}(\boldsymbol{\mu}_t, \boldsymbol{\Sigma}_t)$ at a given timestamp \( t \), it projects the 3D Gaussian ellipsoid $\boldsymbol{x}_t$ into a 2D Gaussian ellipsoid $\boldsymbol{x}_t^{2d}\sim \mathcal{N}(\boldsymbol{\mu}_t^{2d}, \boldsymbol{\Sigma}_t^{2d})$. Therefore, the term \( p_i(u, v, t) \) in Eq.~(\ref{eq:I(uvt)}) can be denoted as:
\vspace{-5px}  
\begin{equation}\label{eq:p_i(u, v, t)}
\begin{aligned}
& p_i(u, v, t)=p\left(u, v ; \boldsymbol{\mu}_{i,t}^{2d}, \boldsymbol{\Sigma}_{i,t}^{2d} \mid t\right) \cdot p(t), \\
\end{aligned}
\vspace{-5px}  
\end{equation}
where \( p(t) \) denotes the time-dependent probability, and 4DGS~\cite{wu20244d} assigns it a default constant value of 1 to adapt to dynamic variations at different time points. Additional details can be found in 4DGS.

Overall, 4DGS merges spatial positioning with temporal information, primarily by incorporating the time variable \( t \) into a probabilistic model to represent dynamic variations, as in the Eq.~(\ref{eq:p_i(u, v, t)}). However, this approach neglects the intricate relationships that exist between dynamic and static scenes, lacking the necessary decoupling of dynamic and static attributes, which significantly complicates dynamic capture and modeling in complex scenarios.

\subsection{Proposed Method: SDD-4DGS}\label{subsec:Separation}
To effectively identify and model the properties of dynamic and static components during 4D scene reconstruction, we propose a probabilistic dynamic perception-based decoupled Gaussian splatting method, as illustrated in Fig.~\ref{fig:pipeline}. This framework achieves static-dynamic separation through the introduction of a dynamic perception coefficient. Below, we elaborate on our approach's theoretical foundation (Sec.~\ref{subsubsec:State Regulation Factor}) and implementation considerations (Sec.~\ref{subsubsec:Uncertainty Guided Decoupling}).

\subsubsection{Static-Dynamic aware Decoupling Framework}\label{subsubsec:State Regulation Factor}


To decouple dynamic and static components within the scene during modeling, we propose the dynamic perception coefficient \( f \) in addition to the timestamp \( t \) for dynamic scene modeling~\cite{wu20244d}. The perception coefficient \( f \)  can be regarded as a learnable parameter added to each Gaussian ellipsoid, which is jointly learned with other reconstruction parameters during optimization.


Specifically, we augment the rendering function in Eq.~(\ref{eq:p_i(u, v, t)}) by introducing a coefficient, enabling the deformed 3D Gaussian projection $p_i(u, v, t)$ to depend on both time \( t \) and a dynamic perception coefficient \( f \):
\vspace{-5px}  
\begin{equation}\label{eq:subsubsec:p_i(u, v, t)}
\begin{aligned}
& p_i(u, v, t)=p\left(u, v ; \boldsymbol{\mu}_{i,t}^{2d}, \boldsymbol{\Sigma}_{i,t}^{2d} \mid t,f_i \right) \cdot p(t, f_i), \\
\end{aligned}
\vspace{-5px}  
\end{equation}
where $p(t, f_i)$ denotes the joint probability distribution of time \( t \) and the dynamic perception coefficient \( f \) of $i$-th Gaussian ellipsoid. Considering the irrelevance between the inherent component properties and time, the joint probability distribution satisfies $p(t, f) = p(t) \cdot p(f)$. Further, based on the work~\cite{wu20244d}, $p(t)$ is assumed to be a constant $1$ for model simplification, thus Eq.~(\ref{eq:subsubsec:p_i(u, v, t)}) is simplified to:
\vspace{-5px}  
\begin{equation}\label{eq:subsubsec:p_i(u, v, t) 2}
\begin{aligned}
& p_i(u, v, t)=p\left(u, v ; \boldsymbol{\mu}_{i,t}^{2d}, \boldsymbol{\Sigma}_{i,t}^{2d} \mid t,f_i\right) \cdot p(f_i). \\
\end{aligned}
\vspace{-5px}  
\end{equation}

Within a given time interval, each Gaussian ellipsoid only exist in only one of two mutually exclusive status, dynamic or static, with \( f_i \) adhering to a Bernoulli distribution, $f \sim \mathcal{B}(w)$, $p(f \mid w)=w^f(1-w)^{1-f}$, which means the $i$-th Gaussian ellipsoid is part of the dynamic scene when \( f_i = 1 \). Consequently, Eq.~(\ref{eq:subsubsec:p_i(u, v, t) 2}) is expressed as:
\vspace{-5px}  
\begin{equation}
\resizebox{.9\hsize}{!}{
$
\begin{aligned}
&p\left(u, v ; \boldsymbol{\mu}_{i,t}^{2d}, \boldsymbol{\Sigma}_{i,t}^{2d} \mid t,f_i\right) p(f_i) \\
 =&\underbrace{p_i\left(u, v \mid t, f_i=1\right)}_{\textit{dynamic characteristic}} p\left(f_i=1\right)+\underbrace{p_i\left(u, v \mid t, f_i=0\right)}_{\textit{static characteristic }} p\left(f_i=0\right),
\end{aligned}\\
$
}
\vspace{-5px}  
\end{equation}
where $p_i\left(u, v \mid t, f_i=1\right)$ and $p_i\left(u, v \mid t, f_i=0\right)$ denote the dynamic and static characteristics of the Gaussian ellipsoid, respectively. Based on 4DGS~\cite{wu20244d}, we characterize the Gaussian’s static attributes within the canonical space as $\boldsymbol{x}_0 \sim \mathcal{N}\left( \boldsymbol{\mu}_0, \boldsymbol{\Sigma}_0 \right)$ and dynamic attributes as $\boldsymbol{x}_t \sim \mathcal{N}\left( \boldsymbol{\mu}_t, \boldsymbol{\Sigma}_t \right)$ in Eq.~(\ref{eq:4dgs Delta}), and represent the deformed Gaussian  $\boldsymbol{x}_t^{'}\sim \mathcal{N}(\boldsymbol{\mu}_t^{'}, \boldsymbol{\Sigma}_t^{'})$ in our framework using the following formulation:
\vspace{-5px}  
\begin{equation}
    \begin{aligned}
        \boldsymbol{\mu}^{'}_{t} 
        = & (1-w) \boldsymbol{\mu}_{0} + w \boldsymbol{\mu}_{t} \\
        = & (1-w) \boldsymbol{\mu}_{0} + w (\boldsymbol{\mu}_{0} + \Delta\boldsymbol{\mu}_{t}) \\
        = & \boldsymbol{\mu}_{0} + w \Delta\boldsymbol{\mu}_{t}, \\
        \boldsymbol{\Sigma}^{'}_{t} 
        = & \boldsymbol{\Sigma}_{0} + w \Delta\boldsymbol{\Sigma}_{t}. \\
    \end{aligned}
    \vspace{-5px}  
\end{equation}

Additionally, by regularizing the probability density of the regulation factor $f$ as follows:
\vspace{-5px}  
\begin{equation}\label{eq:binaryloss}
    \begin{aligned}
        \mathcal{L}_{bi}
        = & - p(f \mid w)log(p(f \mid w)) \\
        = & - (wlog(w)+(1-w)log(1-w)).
    \end{aligned}
    \vspace{-5px}  
\end{equation}
We guide \( p(f) \) towards a single state, meaning each Gaussian ellipsoid possesses only one of the dynamic or static characteristics, further achieving separation of the dynamic component $\mathcal{G}_{d}=\left\{\boldsymbol{\mu}^{'}, \boldsymbol{S}^{'}, \boldsymbol{R}^{'}, c, \alpha, w \mid w > \tau_d \right\}$ and static component $\mathcal{G}_{s}=\left\{\boldsymbol{\mu}^{'}, \boldsymbol{S}^{'}, \boldsymbol{R}^{'}, c, \alpha, w \mid w < \tau_s \right\}$ of the reconstructed scene through dynamic threshold $\tau_t$  and static threshold $\tau_s$ filtering.

\subsubsection{Implementation Optimization Strategy}
\label{subsubsec:Uncertainty Guided Decoupling}
Based on the theoretical derivation of the dynamic perception coefficient proposed in Sec.~\ref{subsubsec:State Regulation Factor}, each Gaussian's dynamic properties can be regulated by the parameter $w$. This parameterization naturally integrates into existing Gaussian reconstruction pipelines enabling adaptive static-dynamic decoupling without requiring complex architectural modifications. However, to fully realize the potential of this theoretical framework, we must address two key implementation challenges.

\textbf{First}, while theoretically the dynamic perception coefficient should converge toward binary states, enforcing this constraint too early in training may degrade the final reconstruction quality, as shown in Tab.~\ref{table:ablation} (the experiments from (\textit{b}) to (\textit{c}) \textit{-0.42dB PSNR}). This occurs because during early training, scene geometry and appearance features have not yet converged, and forcing coefficient binarization may prematurely restrict the model's expressive capacity. To address this issue, we implement a \textit{progressive constraint schedule} through a weighting function
\begin{equation}
    \lambda_{bi}(t) = 1 - e^{-\alpha t}
\end{equation}
, where $\alpha$ controls the rate at which binary constraints are applied. 
This progressive schedule allows the model to first establish fundamental scene geometry before enforcing strict dynamic-static separation.


\begin{figure}[htb]
\vspace{-7px}
\centerline{\includegraphics[width=.95\linewidth]{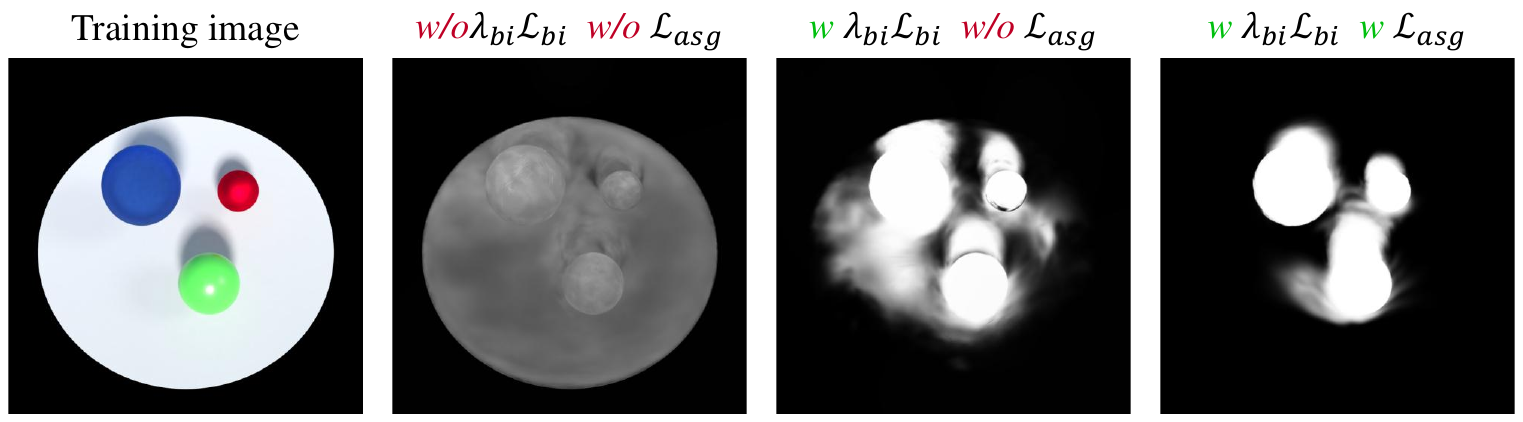}}
    \caption{\textbf{Visualization of optimization strategy effects on dynamic perception coefficient.} 
    From left to right: training image, results without binary constraint and self-supervision (w/o $\lambda_{bi}\mathcal{L}_{bi}$ w/o $\mathcal{L}_{asg}$), with only binary constraint (w $\lambda_{bi}\mathcal{L}_{bi}$ w/o $\mathcal{L}_{asg}$), and with both binary constraint and self-supervision (w $\lambda_{bi}\mathcal{L}_{bi}$ w $\mathcal{L}_{asg}$). The introduction of self-supervision signals significantly improves static-dynamic decoupling.
    }
    \label{fig:mask}
    \vspace{-7px}  
\end{figure}

\textbf{Second}, the core objective of the dynamic perception coefficient is to achieve physically meaningful scene decoupling. However, when optimized solely with reconstruction loss, the model may converge to local optima where stationary backgrounds are incorrectly classified as dynamic regions or slowly moving objects are misidentified as static structures (Fig.~\ref{fig:mask})—errors that directly degrade rendering quality. To address this, we introduce a simple yet effective \textit{automatic supervision signal}. Drawing on the widely observed phenomenon that "dynamic regions typically exhibit higher uncertainty during reconstruction," we incorporate a lightweight uncertainty estimation mechanism~\cite{kulhanek2024wildgaussians} to generate motion masks $m$, enabling end-to-end optimization:
\vspace{-5px}  
\begin{equation}\label{eq:static dynamic loss}
\resizebox{.7\hsize}{!}{$
\begin{aligned}
        \mathcal{L}_{asg} = &(1-m)(\mathcal{L}_1(\hat{I}_s, I) + \mathcal{L}_{ssim}(\hat{I}_s, I)) \\
        & + m(\mathcal{L}_1(\hat{I}_d, I) + \mathcal{L}_{ssim}(\hat{I}_d, I))
\end{aligned}
$}
\vspace{-8px}  
\end{equation}
, where $\hat{I}_d$ and $\hat{I}_s$ represent the renderings from the dynamic part $\mathcal{G}_{d}$ and static part $\mathcal{G}_{s}$ of the scene, respectively.
More implementation details are provided in the supplementary materials.


\subsection{Training Objective}\label{subsec:Optimization}
Our SDD-4DGS framework incorporates a dynamic perception coefficient for effective static-dynamic decoupling. The complete optimization objective consists of three key components:
\begin{equation}
\vspace{-5px}
{\mathcal{L} = \mathcal{L}_{4dgs} + \lambda_{bi}(t)\mathcal{L}_{bi} + \mathcal{L}_{asg}}
\end{equation}
, where $\mathcal{L}_{4dgs}$ represents the fundamental 4DGS reconstruction loss from~\cite{wu20244d}, encompassing rendering fidelity and regularization terms; $\mathcal{L}_{bi}$ denotes the binary entropy loss (Eq.~(\ref{eq:binaryloss})) that constrains the static-dynamic attributes of each Gaussian to converge toward a single state; and $\mathcal{L}_{asg}$ is the automatic supervision guidance loss (Eq.~(\ref{eq:static dynamic loss})) that optimizes dynamic and static regions separately based on uncertainty estimation.

\begin{table*}[htbp]
\caption{\textbf{Quantitative results on the monocular real datasets Nerfies~\cite{park2021nerfies} and HyperNeRF~\cite{park2021hypernerf}.} The \textbf{best} results are highlighted in bold. The rendering resolution is set to $960\times540$.}
\centering
\renewcommand{\arraystretch}{1.2} 
\resizebox{0.8\textwidth}{!}{
\begin{tabular}{lccccccccc}
\Xhline{2pt}
                                     & \multicolumn{2}{c}{\textbf{Broom}}    & \multicolumn{2}{c}{\textbf{3D Printer}} & \multicolumn{2}{c}{\textbf{Chicken}}  & \multicolumn{3}{c}{\textbf{Peel Banana}} \\
\multirow{-2}{*}{\textbf{HyperNeRF}} & PSNR (dB) $\uparrow$ & SSIM $\uparrow$ & PSNR (dB) $\uparrow$  & SSIM $\uparrow$  & PSNR (dB) $\uparrow$ & SSIM $\uparrow$ & PSNR (dB) $\uparrow$  & \multicolumn{2}{c}{SSIM $\uparrow$}  \\ \hline
4DGS~\cite{wu20244d}                                 & 22.40                & 0.641           & 24.55                 & 0.813            & 23.81                & 0.857           & 22.33                 & \multicolumn{2}{c}{0.765}            \\
RealTime4DGS~\cite{yang2023real}                         & 21.25                & 0.560           & 18.04                 & 0.581            & 20.00                & 0.642           & 21.14                 & \multicolumn{2}{c}{0.649}            \\
Spacetime~\cite{li2024spacetime}                            & 21.74                & 0.606           & 22.51                 & 0.753            & 21.16                & 0.793           & 21.72                 & \multicolumn{2}{c}{0.737}            \\
4DRotor~\cite{duan20244d}                              & 20.98                & 0.554           & 18.44                 & 0.630            & 19.09                & 0.694           & 20.09                 & \multicolumn{2}{c}{0.657}            \\
\rowcolor[HTML]{EFEFEF} 
SDD-4DGS (Ours)                                 & \textbf{23.16}       & \textbf{0.673}  & \textbf{24.76}        & \textbf{0.818}   & \textbf{24.67}       & \textbf{0.873}  & \textbf{22.62}        & \multicolumn{2}{c}{\textbf{0.772}}   \\ \Xhline{1pt}
                                     & \multicolumn{2}{c}{\textbf{Curls}}    & \multicolumn{2}{c}{\textbf{Tail}}       & \multicolumn{2}{c}{\textbf{Toby-sit}} & \multicolumn{3}{c}{\textbf{Mean}}        \\
\multirow{-2}{*}{\textbf{Nerfies}}   & PSNR (dB) $\uparrow$ & SSIM $\uparrow$ & PSNR (dB) $\uparrow$  & SSIM $\uparrow$  & PSNR (dB) $\uparrow$ & SSIM $\uparrow$ & PSNR (dB) $\uparrow$  & SSIM $\uparrow$ & FPS $\uparrow$  \\ \hline
4DGS~\cite{wu20244d}                                 & 15.11                & 0.561           & 26.82                 & 0.717            & \textbf{21.45}       & 0.539           & 22.35                 & 0.699   & 60.2        \\
RealTime4DGS~\cite{yang2023real}                         & 17.68                & 0.499           & 23.53                 & 0.555            & 19.87                & 0.384           & 20.21                 & 0.553   & \textbf{72.1}        \\
Spacetime~\cite{li2024spacetime}                            & \textbf{20.26}       & 0.670           & 25.40                 & 0.650            & 21.31                & 0.451           & 22.01                 & 0.665   & 70.2        \\
4DRotor~\cite{duan20244d}                              & 15.50                & 0.546           & 19.89                 & 0.477            & 19.50                & 0.373           & 19.07                 & 0.562   & 38.4        \\
\rowcolor[HTML]{EFEFEF} 
SDD-4DGS (Ours)                                 & 20.14                & \textbf{0.670}  & \textbf{26.67}        & \textbf{0.725}   & 21.23                & \textbf{0.578}  & \textbf{23.32}        & \textbf{0.730}  & 62.3 \\ \Xhline{2pt}
\end{tabular}
}
\label{table:hyper}
\vspace{-10px}  
\end{table*}

\section{Experiment}
In this section, we first detail our implementation and hyperparameter setting in Sec.~\ref{subsec:Experiment-Datasets}. Then, Sec.~\ref{subsec:Experiment-Results} compares our method’s performance against other methods on the five datasets under four distinct setting. Finally, Sec.~\ref{subsec:Experiment-Ablation} provides ablation results, demonstrating the rationale of each module’s design.


\subsection{Implementation $\And$ Hyperparameter Setting}\label{subsec:Experiment-Datasets}
We develop our method based on the PyTorch~\cite{paszke2019pytorch} framework, conducting all experiments on a single RTX 3090 GPU. Specifically, we use the Adam optimizer and trained for \textit{30,000} steps across all datasets.

\begin{figure}[ht]
    \centerline{\includegraphics[width=\linewidth]{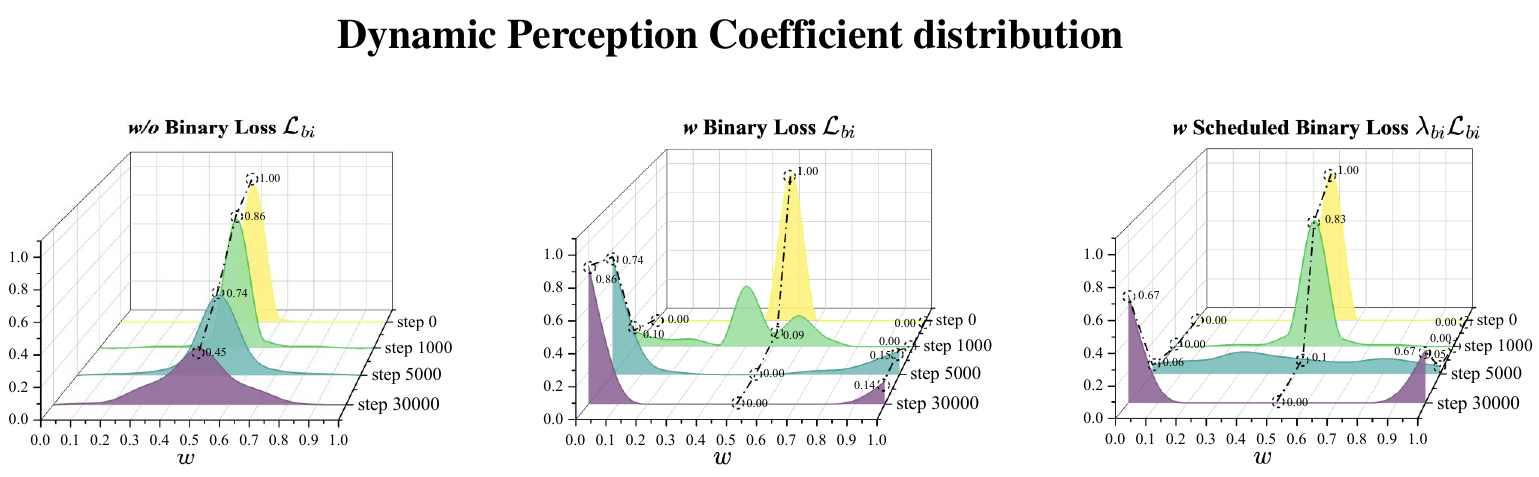}}
    \vspace{-5px}
    \caption{\textbf{Visualization of the dynamic perception coefficient distribution over training steps.} }
    \label{fig:rebuttal_thr}
    \vspace{-10px}  
\end{figure}

We carefully select hyperparameters to balance theoretical alignment with practical performance. For the dynamic perception coefficient thresholds, we empirically determine $\tau_d = \tau_s = 0.5$ during training to ensure smoother optimization, These thresholds optimally separate the emerging bimodal distribution while maintaining reconstruction stability (Fig.~\ref{fig:rebuttal_thr}.). During inference, we tighten these thresholds to $\tau_d = 0.85$ and $\tau_s = 0.2$ to ensure precise component classification after full convergence.
The progressive constraint schedule ($\lambda_{bi}(t) = 1 - e^{-\alpha t}$) represents a critical implementation choice. After extensive experimentation, we set $\alpha = 1 \times 10^{-4}$, which creates an optimal trajectory that allows geometry and appearance to stabilize before enforcing strict dynamic-static classification.
Other parameters, including learning rate, densification, pruning, and opacity reset, are set following prior work~\cite{kerbl20233d}. Further implementation details can be found in the supplementary materials.

\begin{figure*}[ht]
    \centerline{\includegraphics[width=\linewidth]{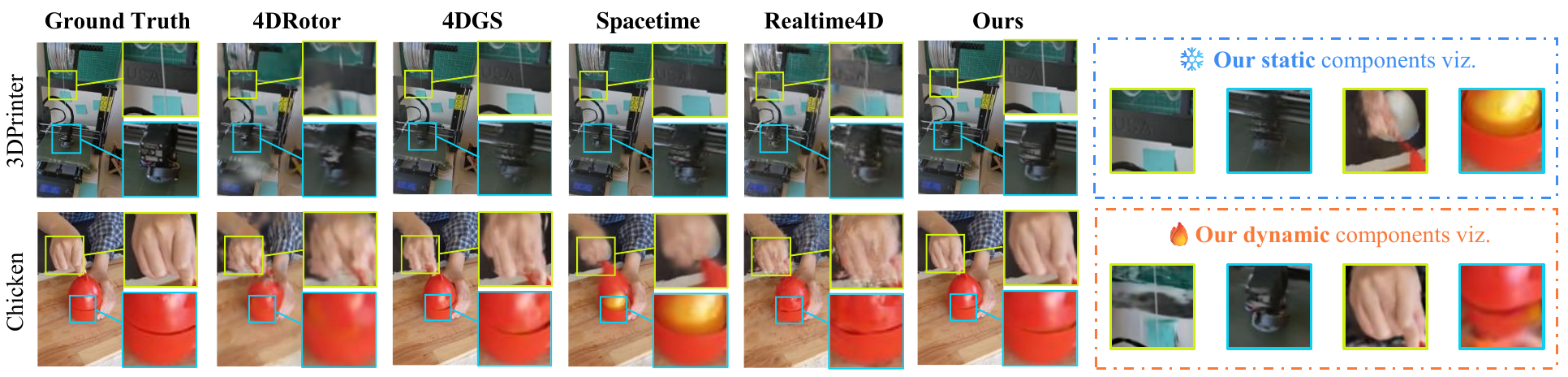}}
    \caption{\textbf{Qualitative comparison between rendered results of HyperNeRF~\cite{park2021hypernerf} dataset.} We visualize the rendering results of our method with those of other methods~\cite{duan20244d, wu20244d, li2024spacetime, yang2023real} and enlarge the local details. In addition, in order to more intuitively demonstrate our separation effect, we render the static and dynamic scenes mentioned in the method separately.}
    \label{fig:hyper}
    \vspace{-15px}  
\end{figure*}

\subsection{Comparison with SOTA}\label{subsec:Experiment-Results}

\noindent\textbf{Evaluation on Monocular Real-world Scene.} Monocular scene reconstruction faces challenges of sparse viewpoints and spatiotemporal imbalances between dynamic and static scene distributions. We test SDD-4DGS and all baselines on HyperNeRF~\cite{park2021hypernerf} and Nerfies~\cite{park2021nerfies}. SDD-4DGS achieves state-of-the-art results in 5 out of 7 scenes (Tab.~\ref{table:hyper}) and improves reconstruction of dynamic textures and high-frequency details (Fig~\ref{fig:hyper}) in all scenes.
Furthermore, we independently render dynamic and static scenes for analysis. SDD-4DGS accurately captures static structures and dynamic movements like printer heads and hand motions. This reduces scene ambiguity, captures static details thoroughly, and integrates dynamic/static information, mitigating issues from viewpoint shifts and sparse observations.



\begin{table}[!htbp]
\caption{
\textbf{Quantitative results on the multi-view real dataset Neu3D’s~\cite{li2022neural}.} The \textbf{best} results are highlighted in bold. The rendering resolution is set to $1352\times1014$.
} 
\vspace{-15px}
\begin{center}
\resizebox{0.42\textwidth}{!}{
\begin{tabular}{lccc}
\Xhline{2pt}
Method                           & PSNR (dB)↑     & SSIM ↑        & LPIPS ↓       \\ \hline
K-Planes~\cite{fridovich2023k}   & 31.63          & -             & -             \\
MixVoxels-X~\cite{wang2023mixed} & 31.73          & -             & \textbf{0.06} \\ \hline
4DGS~\cite{wu20244d}             & 31.23          & 0.93          & 0.12          \\
RealTime4DGS~\cite{yang2023real} & 31.84          & 0.95          & 0.09          \\
Spacetime~\cite{li2024spacetime} & 31.38          & 0.94          & 0.12          \\
4DRotor~\cite{duan20244d}        & 31.73          & 0.94          & 0.10          \\ 
\rowcolor[HTML]{EFEFEF} 
SDD-4DGS (Ours)                             & \textbf{32.39} & \textbf{0.95} & 0.09          \\ \Xhline{2pt}
\end{tabular}
}
\end{center}
\label{table:n3v}
\end{table}

\begin{figure*}[ht]
    \centerline{\includegraphics[width=\linewidth]{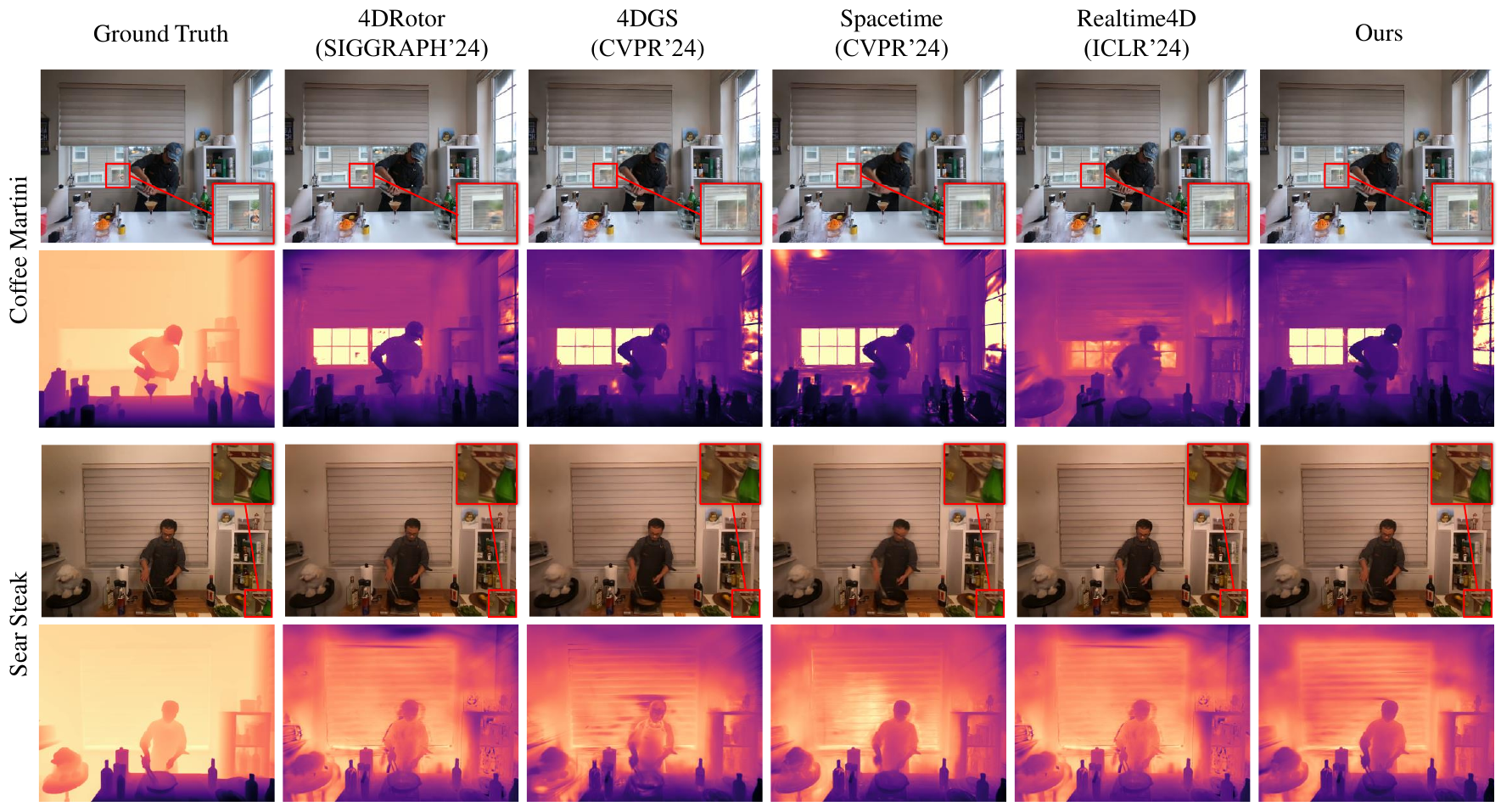}}
    \caption{\textbf{Qualitative comparison between rendered results of Neu3D's~\cite{li2022neural} dataset.} We visualize the rendering results of our method with those of other methods~\cite{duan20244d, wu20244d, li2024spacetime, yang2023real} and enlarge the local details. Additionally, we rendered the depth map of the scene. The ground truth depth map, estimated by method~\cite{yang2024depth}, serves as a reference for comparison. Compared with other methods, the local brightness in our depth map is more consistent.}
    \label{fig:n3v}
    \vspace{-15px}  
\end{figure*}

\noindent\textbf{Evaluation on Multi-view Real-world Scene.} 
As shown in Tab.~\ref{table:n3v}, SDD-4DGS achieves an average PSNR of 32.39dB in Neu3D’s~\cite{li2022neural} dataset, which exceeds the prior best of 31.84dB.
In Fig.~\ref{fig:n3v}, we illustrate the rendering performance and depth of some methods. Most methods achieve satisfactory rendering quality with the scene details captured by multi-view cameras. However, our method demonstrates superior detail in certain aspects, such as objects outside windows, fabric folds, and color rendering of semi-transparent materials. Additionally, in terms of geometric modeling, SDD-4DGS effectively distinguishes foreground from background and achieves enhanced local consistency in depth representation.

\begin{figure}[htbp]
    \centerline{\includegraphics[width=\linewidth]{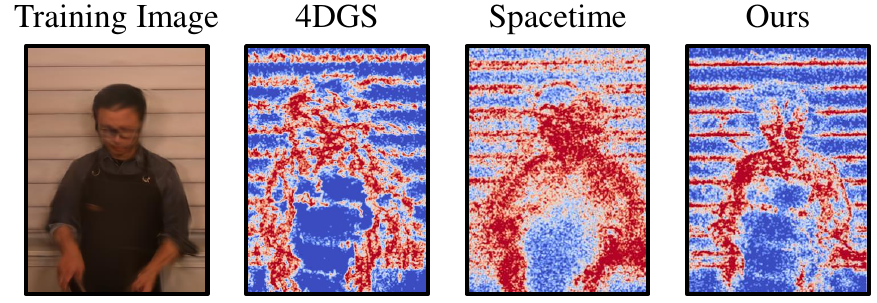}}
    \caption{\textbf{Visualization of point distribution in dynamic regions.} 
     The point density across regions illustrates the distinct spatial distribution achieved by our SDD-4DGS framework in separating static and dynamic components. Compared to prior methods (e.g., 4DGS~\cite{wu20244d} and Spacetime~\cite{li2024spacetime}), our approach demonstrates a more concentrated point allocation which reflects the effectiveness of the proposed decoupling mechanism, reducing static-dynamic interference.}
    \label{fig:point dirtibution}
    \vspace{-10px}  
\end{figure}

To further examine the spatial distribution of points, we provide a visualization in Fig~\ref{fig:point dirtibution}. As observed, our method demonstrates a more focused distribution of points in the dynamic regions, particularly in contrast to 4DGS~\cite{wu20244d} and Spacetime~\cite{li2024spacetime}, where the distribution is affected by static components within the dynamic sections.

\noindent\textbf{Evaluation on Monocular Synthetic Scene.}
As a key benchmark for assessing 4D dynamic scene reconstruction, the monocular synthetic dataset D-NeRF~\cite{pumarola2021d} measures the foundational capabilities of different methods in constructing dynamic scenes. 
To thoroughly evaluate the rendering quality of both dynamic and static components, we manually segmented the dynamic regions within the images, as illustrated in Fig~\ref{fig:intro}(a). We computed the PSNR values for the static regions, dynamic regions, and the overall image separately in Tab.~\ref{table:d-nerf}. Experimental results demonstrate that, due to our decoupling and optimization strategies, SDD-4DGS achieves improved reconstruction quality across both static and dynamic areas, highlighting the critical role of decoupled modeling in enhancing rendering performance.
\begin{table}[!htbp]
\caption{
\textbf{Quantitative results on the synthesis dataset D-NeRF~\cite{pumarola2021d}.} The \textbf{best} results are highlighted in bold. The rendering resolution is set to $800\times800$.
}
\vspace{-15px}  
\begin{center}
\resizebox{0.45\textwidth}{!}{
\begin{tabular}{lcccccc}
\Xhline{2pt}
                             & \multicolumn{3}{c}{\textbf{Trex}}                & \multicolumn{3}{c}{\textbf{Jumping Jacks}}       \\
\multirow{-2}{*}{PSNR (dB)↑} & static         & dynamic        & full           & static         & dynamic        & full           \\ \hline
4DGS~\cite{wu20244d}                         & 25.31          & 20.32          & 25.08          & 34.74          & 25.49          & 34.51          \\
RealTime4DGS~\cite{yang2023real}                 & 25.44          & 19.93          & 24.71          & 29.65          & 21.13          & 29.45          \\
Spacetime~\cite{li2024spacetime}                  & 24.86          & 19.78          & 24.67          & 30.85          & 22.36          & 30.62          \\
4DRotor~\cite{duan20244d}                      & \textbf{25.46} & 22.37          & 25.27          & 29.99          & 25.86          & 29.92          \\
\rowcolor[HTML]{EFEFEF} 
SDD-4DGS (Ours)                          & 25.36          & \textbf{23.38} & \textbf{25.31} & \textbf{35.79} & \textbf{26.98} & \textbf{35.58} \\ \Xhline{1pt}
                             & \multicolumn{3}{c}{\textbf{Mutant}}              & \multicolumn{3}{c}{\textbf{Mean}}                \\
\multirow{-2}{*}{PSNR (dB)↑} & static         & dynamic        & full           & static         & dynamic        & full           \\ \hline
4DGS~\cite{wu20244d}                         & 39.53          & 24.02          & 36.85          & 32.19          & 23.27          & 32.14          \\
RealTime4DGS~\cite{yang2023real}                 & 33.55          & 22.65          & 32.43          & 29.54          & 21.23          & 28.86          \\
Spacetime~\cite{li2024spacetime}                  & 34.94          & 24.97          & 32.82          & 30.21          & 22.37          & 29.37          \\
4DRotor~\cite{duan20244d}                      & 39.44          & \textbf{27.85}          & 38.15          & 31.63          & 25.36          & 31.11          \\
\rowcolor[HTML]{EFEFEF} 
SDD-4DGS (Ours)                         & \textbf{40.35} & 27.42 & \textbf{38.67} & \textbf{33.83} & \textbf{25.92} & \textbf{33.18} \\ \Xhline{2pt}
\end{tabular}
}
\end{center}
\label{table:d-nerf}

\end{table}

\begin{figure}[htbp]
    \centerline{\includegraphics[width=\linewidth]{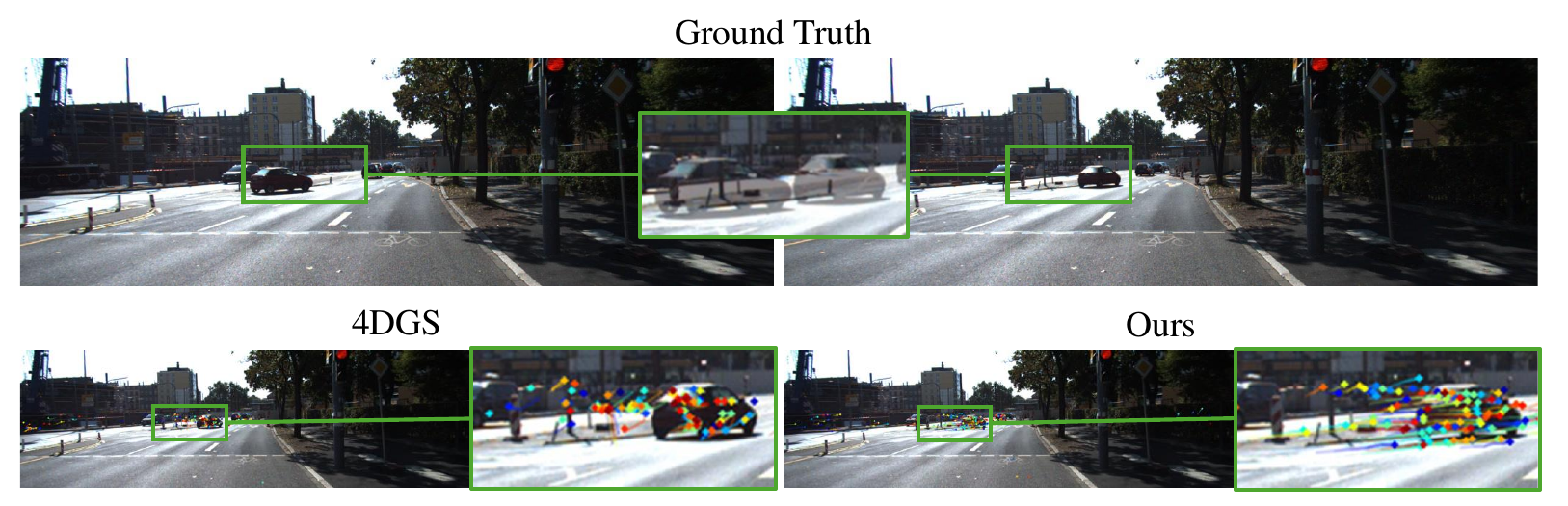}}
    \caption{\textbf{Visualization of tracking with 3D Gaussians.} 
    We superimpose local pictures of the ground truth to show the real motion situation and visualize the motion trajectory of the 3D Gausssians predicted by 4DGS~\cite{wu20244d} and our method.}
    \label{fig:tracking}
    \vspace{-15px}  
\end{figure}
\noindent\textbf{Urban Scene Evaluation.} High-speed vehicle traffic in urban environments challenges 4D reconstruction methodologies. Our approach employs targeted supervision for dynamic regions during training, enhancing dynamic capture. Fig.~\ref{fig:tracking} shows sampled points and their tracking signals, highlighting our method's superior motion modeling for fast-moving objects over 4DGS~\cite{wu20244d}. Quantitative results are detailed in the supplementary materials.

\subsection{Ablation Studies}\label{subsec:Experiment-Ablation}
Table~\ref{table:ablation} demonstrates the impact of our method's core components on reconstruction performance using the D-NeRF~\cite{pumarola2021d} dataset. We systematically evaluated our implementation strategies through controlled experiments to analyze the effects of progressive binary constraint and automatic supervision signals.

\noindent\textbf{Dynamic Perception Coefficient Analysis:} Our baseline configuration (Tab.~\ref{table:ablation} (\textit{a})) represents the standard 4D Gaussian Splatting approach~\cite{wu20244d} without our proposed components, achieving a PSNR of 34.14dB. Introducing the dynamic perception coefficient $w$ (Tab.~\ref{table:ablation} (\textit{b})) yields a notable improvement to 34.34dB PSNR and 0.95 SSIM. This improvement validates our theoretical assertion that explicitly modeling the static-dynamic nature of scene components through a probabilistic framework enhances reconstruction fidelity.

\noindent\textbf{Effect of Progressive Binary Constraint:} Tab.~\ref{table:ablation} (\textit{c}) shows that introducing binary entropy loss ($\mathcal{L}_{bi}$) without progressive weight scheduling ($\lambda_{bi}$) results in a PSNR decrease. However, when the progressive weight scheduling mechanism is incorporated (Tab.~\ref{table:ablation} (\textit{c})), performance significantly improves to 34.63dB. This confirms our argument in Section~\ref{subsubsec:Uncertainty Guided Decoupling} that enforcing binarization too early restricts model expressivity, while progressive constraint introduction effectively balances reconstruction quality and static-dynamic separation.

\noindent\textbf{Contribution of Automatic Supervision:} The experiments from (\textit{d}) to (\textit{Full}) in Tab.~\ref{table:ablation} demonstrate the value of automatic supervision signals. When introducing the explicit loss $\mathcal{L}_{asg}$, PSNR improves by 0.19dB, and it enables the model to achieve optimal performance. This progression confirms that automatic supervision signals effectively address local optima issues in dynamic-static classification, particularly in correctly separating complex scenes.

\begin{table}[htbp]
\caption{\textbf{Ablation study on the D-NeRF dataset.}
The baseline (\textit{a}) represents standard 4DGS~\cite{wu20244d}. The dynamic perception coefficient $w$ (\textit{b}) and the combination of binary entropy loss $\mathcal{L}_{bi}$ with progressive scheduling ($\lambda_{bi}L_{bi}$) (\textit{d}) improve performance. The \textit{Full} model, which includes automatic supervision signals $\mathcal{L}_{asg}$, achieves optimal results, highlighting the effectiveness of our static-dynamic aware decoupling framework.
}
\begin{center}
\resizebox{0.4\textwidth}{!}{
\begin{tabular}{lcccc|cc}
\Xhline{2pt}
                     & \multicolumn{4}{c|}{\textbf{Ablation Items}}                                                                & \multicolumn{2}{c}{\textbf{D-NeRF}}                     \\
\multirow{-2}{*}{ID} & $w$ & $\mathcal{L}_{bi}$                    & $\lambda_{bi}$                  & $\mathcal{L}_{asg}$                                      & PSNR (dB)$\uparrow$     & SSIM $\uparrow$                     \\ \hline
\textit{a}           &                           &                           &                           &                           & 34.14          & 0.94                      \\
\textit{b}           & $\checkmark$                &                           &                           &                           & 34.34          & 0.95                      \\
\textit{c}           & $\checkmark$                & $\checkmark$                &                           &                           & 33.92          & 0.94                      \\
\textit{d}           & $\checkmark$                & $\checkmark$                & $\checkmark$                &                           & 34.63          & 0.95                      \\
\rowcolor[HTML]{EFEFEF} 
\textit{Full}        & $\checkmark$                & $\checkmark$                & $\checkmark$                & $\checkmark$                & \textbf{34.82}          & \textbf{0.96}                      \\ \Xhline{2pt}
\end{tabular}
}
\end{center}
\label{table:ablation}
\end{table}

\section{Conclusion}
We propose SDD-4DGS, an innovative framework for 4D scene reconstruction that employs static-dynamic aware decoupling within Gaussian Splatting. Our method incorporates a probabilistic formulation of dynamic perception into the Gaussian reconstruction pipeline. With optimized strategies, we leverage this framework in practical applications. Evaluations on various datasets show our approach outperforms existing techniques, reducing scene ambiguity and static-dynamic interference. This framework provides a robust solution for real-world 4D reconstructions, with potential for future research in complex scenarios like dynamic lighting and deformable objects, as well as integration with neural rendering for real-time use.

{
    \small
    \bibliographystyle{ieeenat_fullname}
    \bibliography{main}

\begin{thebibliography}{47}
\providecommand{\natexlab}[1]{#1}
\providecommand{\url}[1]{\texttt{#1}}
\expandafter\ifx\csname urlstyle\endcsname\relax
  \providecommand{\doi}[1]{doi: #1}\else
  \providecommand{\doi}{doi: \begingroup \urlstyle{rm}\Url}\fi

\bibitem[Alfonso-Arsuaga et~al.(2024)Alfonso-Arsuaga, Garc{\'\i}a-Gonz{\'a}lez, Castiella-Aguirrezabala, Alonso, and Garc{\'e}s]{alfonso2024dynerfactor}
Mario Alfonso-Arsuaga, Jorge Garc{\'\i}a-Gonz{\'a}lez, Andrea Castiella-Aguirrezabala, Miguel~Andr{\'e}s Alonso, and Elena Garc{\'e}s.
\newblock Dynerfactor: Temporally consistent intrinsic scene decomposition for dynamic nerfs.
\newblock \emph{Computers \& Graphics}, 122:\penalty0 103984, 2024.

\bibitem[Cao and Johnson(2023)]{cao2023hexplane}
Ang Cao and Justin Johnson.
\newblock Hexplane: A fast representation for dynamic scenes.
\newblock In \emph{Proceedings of the IEEE/CVF Conference on Computer Vision and Pattern Recognition}, pages 130--141, 2023.

\bibitem[Chen and Tsukada(2022)]{chen2022flow}
Quei-An Chen and Akihiro Tsukada.
\newblock Flow supervised neural radiance fields for static-dynamic decomposition.
\newblock In \emph{2022 International Conference on Robotics and Automation (ICRA)}, pages 10641--10647. IEEE, 2022.

\bibitem[Cordts et~al.(2016)Cordts, Omran, Ramos, Rehfeld, Enzweiler, Benenson, Franke, Roth, and Schiele]{cordts2016cityscapes}
Marius Cordts, Mohamed Omran, Sebastian Ramos, Timo Rehfeld, Markus Enzweiler, Rodrigo Benenson, Uwe Franke, Stefan Roth, and Bernt Schiele.
\newblock The cityscapes dataset for semantic urban scene understanding.
\newblock In \emph{Proceedings of the IEEE conference on computer vision and pattern recognition}, pages 3213--3223, 2016.

\bibitem[Dong et~al.(2018)Dong, Shi, Tang, Wang, and Zha]{dong2018efficient}
Wei Dong, Jieqi Shi, Weijie Tang, Xin Wang, and Hongbin Zha.
\newblock An efficient volumetric mesh representation for real-time scene reconstruction using spatial hashing.
\newblock In \emph{2018 IEEE International Conference on Robotics and Automation (ICRA)}, pages 6323--6330. IEEE, 2018.

\bibitem[Duan et~al.(2024)Duan, Wei, Dai, He, Chen, and Chen]{duan20244d}
Yuanxing Duan, Fangyin Wei, Qiyu Dai, Yuhang He, Wenzheng Chen, and Baoquan Chen.
\newblock 4d-rotor gaussian splatting: towards efficient novel view synthesis for dynamic scenes.
\newblock In \emph{ACM SIGGRAPH 2024 Conference Papers}, pages 1--11, 2024.

\bibitem[Fridovich-Keil et~al.(2023)Fridovich-Keil, Meanti, Warburg, Recht, and Kanazawa]{fridovich2023k}
Sara Fridovich-Keil, Giacomo Meanti, Frederik~Rahb{\ae}k Warburg, Benjamin Recht, and Angjoo Kanazawa.
\newblock K-planes: Explicit radiance fields in space, time, and appearance.
\newblock In \emph{Proceedings of the IEEE/CVF Conference on Computer Vision and Pattern Recognition}, pages 12479--12488, 2023.

\bibitem[Gan et~al.(2023)Gan, Xu, Huang, Chen, and Yokoya]{gan2023v4d}
Wanshui Gan, Hongbin Xu, Yi Huang, Shifeng Chen, and Naoto Yokoya.
\newblock V4d: Voxel for 4d novel view synthesis.
\newblock \emph{IEEE Transactions on Visualization and Computer Graphics}, 2023.

\bibitem[Gao et~al.(2021)Gao, Saraf, Kopf, and Huang]{gao2021dynamic}
Chen Gao, Ayush Saraf, Johannes Kopf, and Jia-Bin Huang.
\newblock Dynamic view synthesis from dynamic monocular video.
\newblock In \emph{Proceedings of the IEEE/CVF International Conference on Computer Vision}, pages 5712--5721, 2021.

\bibitem[Gotardo et~al.(2015)Gotardo, Simon, Sheikh, and Matthews]{gotardo2015photogeometric}
Paulo~FU Gotardo, Tomas Simon, Yaser Sheikh, and Iain Matthews.
\newblock Photogeometric scene flow for high-detail dynamic 3d reconstruction.
\newblock In \emph{Proceedings of the IEEE international conference on computer vision}, pages 846--854, 2015.

\bibitem[Huang et~al.(2022)Huang, Gojcic, Huang, Wieser, and Schindler]{huang2022dynamic}
Shengyu Huang, Zan Gojcic, Jiahui Huang, Andreas Wieser, and Konrad Schindler.
\newblock Dynamic 3d scene analysis by point cloud accumulation.
\newblock In \emph{European Conference on Computer Vision}, pages 674--690. Springer, 2022.

\bibitem[Ingale et~al.(2021)]{ingale2021real}
Anupama~K Ingale et~al.
\newblock Real-time 3d reconstruction techniques applied in dynamic scenes: A systematic literature review.
\newblock \emph{Computer Science Review}, 39:\penalty0 100338, 2021.

\bibitem[Kaveti et~al.(2020)Kaveti, Katt, and Singh]{kaveti2020removing}
Pushyami Kaveti, Sammie Katt, and Hanumant Singh.
\newblock Removing dynamic objects for static scene reconstruction using light fields.
\newblock \emph{arXiv preprint arXiv:2003.11076}, 2020.

\bibitem[Kerbl et~al.(2023)Kerbl, Kopanas, Leimk{\"u}hler, and Drettakis]{kerbl20233d}
Bernhard Kerbl, Georgios Kopanas, Thomas Leimk{\"u}hler, and George Drettakis.
\newblock 3d gaussian splatting for real-time radiance field rendering.
\newblock \emph{ACM Trans. Graph.}, 42\penalty0 (4):\penalty0 139--1, 2023.

\bibitem[Kim et~al.(2012)Kim, Guillemaut, Takai, Sarim, and Hilton]{kim2012outdoor}
Hansung Kim, Jean-Yves Guillemaut, Takeshi Takai, Muhammad Sarim, and Adrian Hilton.
\newblock Outdoor dynamic 3-d scene reconstruction.
\newblock \emph{IEEE Transactions on Circuits and Systems for Video Technology}, 22\penalty0 (11):\penalty0 1611--1622, 2012.

\bibitem[Kulhanek et~al.(2024)Kulhanek, Peng, Kukelova, Pollefeys, and Sattler]{kulhanek2024wildgaussians}
Jonas Kulhanek, Songyou Peng, Zuzana Kukelova, Marc Pollefeys, and Torsten Sattler.
\newblock Wildgaussians: 3d gaussian splatting in the wild.
\newblock \emph{arXiv preprint arXiv:2407.08447}, 2024.

\bibitem[Li et~al.(2022)Li, Slavcheva, Zollhoefer, Green, Lassner, Kim, Schmidt, Lovegrove, Goesele, Newcombe, et~al.]{li2022neural}
Tianye Li, Mira Slavcheva, Michael Zollhoefer, Simon Green, Christoph Lassner, Changil Kim, Tanner Schmidt, Steven Lovegrove, Michael Goesele, Richard Newcombe, et~al.
\newblock Neural 3d video synthesis from multi-view video.
\newblock In \emph{Proceedings of the IEEE/CVF Conference on Computer Vision and Pattern Recognition}, pages 5521--5531, 2022.

\bibitem[Li et~al.(2021)Li, Niklaus, Snavely, and Wang]{li2021neural}
Zhengqi Li, Simon Niklaus, Noah Snavely, and Oliver Wang.
\newblock Neural scene flow fields for space-time view synthesis of dynamic scenes.
\newblock In \emph{Proceedings of the IEEE/CVF Conference on Computer Vision and Pattern Recognition}, pages 6498--6508, 2021.

\bibitem[Li et~al.(2024)Li, Chen, Li, and Xu]{li2024spacetime}
Zhan Li, Zhang Chen, Zhong Li, and Yi Xu.
\newblock Spacetime gaussian feature splatting for real-time dynamic view synthesis.
\newblock In \emph{Proceedings of the IEEE/CVF Conference on Computer Vision and Pattern Recognition}, pages 8508--8520, 2024.

\bibitem[Liu et~al.(2020)Liu, Gu, Zaw~Lin, Chua, and Theobalt]{liu2020neural}
Lingjie Liu, Jiatao Gu, Kyaw Zaw~Lin, Tat-Seng Chua, and Christian Theobalt.
\newblock Neural sparse voxel fields.
\newblock \emph{Advances in Neural Information Processing Systems}, 33:\penalty0 15651--15663, 2020.

\bibitem[Liu et~al.(2019)Liu, Yan, and Bohg]{liu2019meteornet}
Xingyu Liu, Mengyuan Yan, and Jeannette Bohg.
\newblock Meteornet: Deep learning on dynamic 3d point cloud sequences.
\newblock In \emph{Proceedings of the IEEE/CVF International Conference on Computer Vision}, pages 9246--9255, 2019.

\bibitem[Liu et~al.(2023)Liu, Gao, Meuleman, Tseng, Saraf, Kim, Chuang, Kopf, and Huang]{liu2023robust}
Yu-Lun Liu, Chen Gao, Andreas Meuleman, Hung-Yu Tseng, Ayush Saraf, Changil Kim, Yung-Yu Chuang, Johannes Kopf, and Jia-Bin Huang.
\newblock Robust dynamic radiance fields.
\newblock In \emph{Proceedings of the IEEE/CVF Conference on Computer Vision and Pattern Recognition}, pages 13--23, 2023.

\bibitem[Lu et~al.(2024)Lu, Wang, Fan, Lu, Li, and Tang]{lu2024image}
Yujie Lu, Shuo Wang, Sensen Fan, Jiahui Lu, Peixian Li, and Pingbo Tang.
\newblock Image-based 3d reconstruction for multi-scale civil and infrastructure projects: A review from 2012 to 2022 with new perspective from deep learning methods.
\newblock \emph{Advanced Engineering Informatics}, 59:\penalty0 102268, 2024.

\bibitem[Menze and Geiger(2015)]{menze2015object}
Moritz Menze and Andreas Geiger.
\newblock Object scene flow for autonomous vehicles.
\newblock In \emph{Proceedings of the IEEE conference on computer vision and pattern recognition}, pages 3061--3070, 2015.

\bibitem[Mildenhall et~al.(2021)Mildenhall, Srinivasan, Tancik, Barron, Ramamoorthi, and Ng]{mildenhall2021nerf}
Ben Mildenhall, Pratul~P Srinivasan, Matthew Tancik, Jonathan~T Barron, Ravi Ramamoorthi, and Ren Ng.
\newblock Nerf: Representing scenes as neural radiance fields for view synthesis.
\newblock \emph{Communications of the ACM}, 65\penalty0 (1):\penalty0 99--106, 2021.

\bibitem[Oquab et~al.(2023)Oquab, Darcet, Moutakanni, Vo, Szafraniec, Khalidov, Fernandez, Haziza, Massa, El-Nouby, et~al.]{oquab2023dinov2}
Maxime Oquab, Timoth{\'e}e Darcet, Th{\'e}o Moutakanni, Huy Vo, Marc Szafraniec, Vasil Khalidov, Pierre Fernandez, Daniel Haziza, Francisco Massa, Alaaeldin El-Nouby, et~al.
\newblock Dinov2: Learning robust visual features without supervision.
\newblock \emph{arXiv preprint arXiv:2304.07193}, 2023.

\bibitem[Ost et~al.(2021)Ost, Mannan, Thuerey, Knodt, and Heide]{ost2021neural}
Julian Ost, Fahim Mannan, Nils Thuerey, Julian Knodt, and Felix Heide.
\newblock Neural scene graphs for dynamic scenes.
\newblock In \emph{Proceedings of the IEEE/CVF Conference on Computer Vision and Pattern Recognition}, pages 2856--2865, 2021.

\bibitem[Park et~al.(2021{\natexlab{a}})Park, Sinha, Barron, Bouaziz, Goldman, Seitz, and Martin-Brualla]{park2021nerfies}
Keunhong Park, Utkarsh Sinha, Jonathan~T Barron, Sofien Bouaziz, Dan~B Goldman, Steven~M Seitz, and Ricardo Martin-Brualla.
\newblock Nerfies: Deformable neural radiance fields.
\newblock In \emph{Proceedings of the IEEE/CVF International Conference on Computer Vision}, pages 5865--5874, 2021{\natexlab{a}}.

\bibitem[Park et~al.(2021{\natexlab{b}})Park, Sinha, Hedman, Barron, Bouaziz, Goldman, Martin-Brualla, and Seitz]{park2021hypernerf}
Keunhong Park, Utkarsh Sinha, Peter Hedman, Jonathan~T Barron, Sofien Bouaziz, Dan~B Goldman, Ricardo Martin-Brualla, and Steven~M Seitz.
\newblock Hypernerf: A higher-dimensional representation for topologically varying neural radiance fields.
\newblock \emph{arXiv preprint arXiv:2106.13228}, 2021{\natexlab{b}}.

\bibitem[Paszke et~al.(2019)Paszke, Gross, Massa, Lerer, Bradbury, Chanan, Killeen, Lin, Gimelshein, Antiga, et~al.]{paszke2019pytorch}
Adam Paszke, Sam Gross, Francisco Massa, Adam Lerer, James Bradbury, Gregory Chanan, Trevor Killeen, Zeming Lin, Natalia Gimelshein, Luca Antiga, et~al.
\newblock Pytorch: An imperative style, high-performance deep learning library.
\newblock \emph{Advances in neural information processing systems}, 32, 2019.

\bibitem[Pumarola et~al.(2021)Pumarola, Corona, Pons-Moll, and Moreno-Noguer]{pumarola2021d}
Albert Pumarola, Enric Corona, Gerard Pons-Moll, and Francesc Moreno-Noguer.
\newblock D-nerf: Neural radiance fields for dynamic scenes.
\newblock In \emph{Proceedings of the IEEE/CVF Conference on Computer Vision and Pattern Recognition}, pages 10318--10327, 2021.

\bibitem[Ren et~al.(2024)Ren, Zhu, Sun, Chen, Pollefeys, and Peng]{ren2024nerf}
Weining Ren, Zihan Zhu, Boyang Sun, Jiaqi Chen, Marc Pollefeys, and Songyou Peng.
\newblock Nerf on-the-go: Exploiting uncertainty for distractor-free nerfs in the wild.
\newblock In \emph{Proceedings of the IEEE/CVF Conference on Computer Vision and Pattern Recognition}, pages 8931--8940, 2024.

\bibitem[Rudnev et~al.(2022)Rudnev, Elgharib, Smith, Liu, Golyanik, and Theobalt]{rudnev2022nerf}
Viktor Rudnev, Mohamed Elgharib, William Smith, Lingjie Liu, Vladislav Golyanik, and Christian Theobalt.
\newblock Nerf for outdoor scene relighting.
\newblock In \emph{European Conference on Computer Vision}, pages 615--631. Springer, 2022.

\bibitem[Tewari et~al.(2022)Tewari, Thies, Mildenhall, Srinivasan, Tretschk, Yifan, Lassner, Sitzmann, Martin-Brualla, Lombardi, et~al.]{tewari2022advances}
Ayush Tewari, Justus Thies, Ben Mildenhall, Pratul Srinivasan, Edgar Tretschk, Wang Yifan, Christoph Lassner, Vincent Sitzmann, Ricardo Martin-Brualla, Stephen Lombardi, et~al.
\newblock Advances in neural rendering.
\newblock In \emph{Computer Graphics Forum}, pages 703--735. Wiley Online Library, 2022.

\bibitem[Tschernezki et~al.(2021)Tschernezki, Larlus, and Vedaldi]{tschernezki2021neuraldiff}
Vadim Tschernezki, Diane Larlus, and Andrea Vedaldi.
\newblock Neuraldiff: Segmenting 3d objects that move in egocentric videos.
\newblock In \emph{2021 International Conference on 3D Vision (3DV)}, pages 910--919. IEEE, 2021.

\bibitem[Wang et~al.(2023)Wang, Tan, Li, Tian, Song, and Liu]{wang2023mixed}
Feng Wang, Sinan Tan, Xinghang Li, Zeyue Tian, Yafei Song, and Huaping Liu.
\newblock Mixed neural voxels for fast multi-view video synthesis.
\newblock In \emph{Proceedings of the IEEE/CVF International Conference on Computer Vision}, pages 19706--19716, 2023.

\bibitem[Wu et~al.(2024)Wu, Yi, Fang, Xie, Zhang, Wei, Liu, Tian, and Wang]{wu20244d}
Guanjun Wu, Taoran Yi, Jiemin Fang, Lingxi Xie, Xiaopeng Zhang, Wei Wei, Wenyu Liu, Qi Tian, and Xinggang Wang.
\newblock 4d gaussian splatting for real-time dynamic scene rendering.
\newblock In \emph{Proceedings of the IEEE/CVF Conference on Computer Vision and Pattern Recognition}, pages 20310--20320, 2024.

\bibitem[Wu et~al.(2022)Wu, Zhong, Tagliasacchi, Cole, and Oztireli]{wu2022d}
Tianhao Wu, Fangcheng Zhong, Andrea Tagliasacchi, Forrester Cole, and Cengiz Oztireli.
\newblock D\^{} 2nerf: Self-supervised decoupling of dynamic and static objects from a monocular video.
\newblock \emph{Advances in neural information processing systems}, 35:\penalty0 32653--32666, 2022.

\bibitem[Xie et~al.(2023)Xie, Guo, Li, Liu, and Xu]{xie2023deform2nerf}
Xiaolong Xie, Xusheng Guo, Wei Li, Jie Liu, and Jianfeng Xu.
\newblock Deform2nerf: Non-rigid deformation and 2d--3d feature fusion with cross-attention for dynamic human reconstruction.
\newblock \emph{Electronics}, 12\penalty0 (21):\penalty0 4382, 2023.

\bibitem[Yan et~al.(2023)Yan, Li, and Lee]{yan2023nerf}
Zhiwen Yan, Chen Li, and Gim~Hee Lee.
\newblock Nerf-ds: Neural radiance fields for dynamic specular objects.
\newblock In \emph{Proceedings of the IEEE/CVF Conference on Computer Vision and Pattern Recognition}, pages 8285--8295, 2023.

\bibitem[Yang et~al.(2023{\natexlab{a}})Yang, Yang, Zhang, Manchester, and Ramanan]{yang2023ppr}
Gengshan Yang, Shuo Yang, John~Z Zhang, Zachary Manchester, and Deva Ramanan.
\newblock Ppr: Physically plausible reconstruction from monocular videos.
\newblock In \emph{Proceedings of the IEEE/CVF International Conference on Computer Vision}, pages 3914--3924, 2023{\natexlab{a}}.

\bibitem[Yang et~al.(2024)Yang, Kang, Huang, Zhao, Xu, Feng, and Zhao]{yang2024depth}
Lihe Yang, Bingyi Kang, Zilong Huang, Zhen Zhao, Xiaogang Xu, Jiashi Feng, and Hengshuang Zhao.
\newblock Depth anything v2.
\newblock \emph{arXiv preprint arXiv:2406.09414}, 2024.

\bibitem[Yang et~al.(2023{\natexlab{b}})Yang, Yang, Pan, and Zhang]{yang2023real}
Zeyu Yang, Hongye Yang, Zijie Pan, and Li Zhang.
\newblock Real-time photorealistic dynamic scene representation and rendering with 4d gaussian splatting.
\newblock \emph{arXiv preprint arXiv:2310.10642}, 2023{\natexlab{b}}.

\bibitem[Yunus et~al.(2024)Yunus, Lenssen, Niemeyer, Liao, Rupprecht, Theobalt, Pons-Moll, Huang, Golyanik, and Ilg]{yunus2024recent}
Raza Yunus, Jan~Eric Lenssen, Michael Niemeyer, Yiyi Liao, Christian Rupprecht, Christian Theobalt, Gerard Pons-Moll, Jia-Bin Huang, Vladislav Golyanik, and Eddy Ilg.
\newblock Recent trends in 3d reconstruction of general non-rigid scenes.
\newblock In \emph{Computer Graphics Forum}, page e15062. Wiley Online Library, 2024.

\bibitem[Zhang et~al.(2021)Zhang, Bengio, Hardt, Recht, and Vinyals]{zhang2021understanding}
Chiyuan Zhang, Samy Bengio, Moritz Hardt, Benjamin Recht, and Oriol Vinyals.
\newblock Understanding deep learning (still) requires rethinking generalization.
\newblock \emph{Communications of the ACM}, 64\penalty0 (3):\penalty0 107--115, 2021.

\bibitem[Zhang and Xu(2017)]{zhang2017mixedfusion}
Hao Zhang and Feng Xu.
\newblock Mixedfusion: Real-time reconstruction of an indoor scene with dynamic objects.
\newblock \emph{IEEE transactions on visualization and computer graphics}, 24\penalty0 (12):\penalty0 3137--3146, 2017.

\bibitem[Zhu et~al.(2023)Zhu, Wan, Tang, and Shi]{zhu2023occlusion}
Chengxuan Zhu, Renjie Wan, Yunkai Tang, and Boxin Shi.
\newblock Occlusion-free scene recovery via neural radiance fields.
\newblock In \emph{Proceedings of the IEEE/CVF Conference on Computer Vision and Pattern Recognition}, pages 20722--20731, 2023.

\end{thebibliography}
}

\clearpage
\setcounter{page}{1}
\maketitlesupplementary

\section{Dataset Description}

This section primarily presents detailed information about the selected datasets, which are categorized into four different types: monocular synthetic datasets (D-NeRF~\cite{pumarola2021d}), monocular real datasets (HyperNeRF~\cite{park2021hypernerf} \& Nerfies~\cite{park2021nerfies}), multi-view real datasets (Neu3D's~\cite{li2022neural}), and real street scene datasets (KITTI~\cite{menze2015object}). Detailed dataset information can be found in Tab.~\ref{table:Datasets}, followed by descriptions of the data partitioning and related explanations.

\noindent\textbf{D-NeRF}~\cite{pumarola2021d} dataset comprises 8 scenes, with default training and testing splits. The highest resolution of $800 \times 800$ is selected for rendering.
comparisons.

\noindent\textbf{HyperNeRF}~\cite{park2021hypernerf} dataset contains scenes from 14 monocular cameras and 4 binocular cameras. We select the four binocular scenes(\textit{vrig-peel-banana, vrig-chiken, vrig-3dprinter, broom2}) for novel view synthesis. One camera is used for training and another for validation. We downsample to a resolution of 960x540 for rendering and initialize the point cloud using COLMAP.
\begin{table}[b]
\vspace{-10px}
\caption{\textbf{Dataset overview used in the analysis.}
The table lists the number of scenes, the presence of point clouds, and rendering resolutions. Further details on data partitioning are discussed in the following sections.}
\vspace{-10px}
\begin{center}
\resizebox{0.45\textwidth}{!}{
\begin{tabular}{lccccc}
\Xhline{2pt}
\textbf{Datasets} & \textbf{\begin{tabular}[c]{@{}c@{}}Camera\\ View\end{tabular}} & \textbf{\begin{tabular}[c]{@{}c@{}}Scene \\ Category\end{tabular}} & \textbf{\begin{tabular}[c]{@{}c@{}}Scenes\\ Num.\end{tabular}} & \textbf{\begin{tabular}[c]{@{}c@{}}Initial \\ Point Cloud\end{tabular}} & \textbf{\begin{tabular}[c]{@{}c@{}}Resolution\end{tabular}} \\ 
\midrule
D-NeRF~\cite{pumarola2021d}            & Monocular                                                           & Synthetic                                                    & 8                                                              & -                                                              & 800 $\times$ 800                          \\ 
HyperNeRF~\cite{park2021hypernerf}         & Monocular                                                           & Outdoor                                                         & 4                                                              & COLMAP                                                              & 960 $\times$  540                        \\ 
Nerfies~\cite{park2021nerfies}           & Monocular                                                           & Outdoor                                                         & 4                                                              & COLMAP                                                              & 960 $\times$  540                        \\ 
Neu3D'S~\cite{li2022neural}               & Multi-view                                                           & Indoor                                                        & 5                                                              & COLMAP                                                             & 1352 $\times$ 1014                        \\ 
KITTI~\cite{menze2015object}             & Monocular                                                           & Urban                                                 & 7                                                              & COLMAP                                                             & 1242 $\times$ 375                        \\ 

\Xhline{2pt}
\end{tabular}}
\end{center}
\label{table:Datasets}
\end{table}

\noindent\textbf{Nerfies}~\cite{park2021nerfies} dataset includes 4 binocular scenes, with \textit{"broom"} being very similar to \textit{"broom2"} in HyperNerf dataset. Therefore, we select three scenes (\textit{curls, tail toby-sit}) for experiments. Similarly to the HyperNerf setup, we designate one camera for training and use the remaining camera for validation. We downsample to a resolution of 960x540 for rendering and initialize the point cloud using COLMAP.

\noindent\textbf{Neu3D's}~\cite{li2022neural} dataset includes six multi-view scenes captured with 20 high-definition cameras in an indoor kitchen. We select five frequently used scenes for comparison (excluding flame\_salmon). The original 2K video is downsampled to $1352 \times 1014$ for rendering, and we use the provided point clouds for initialization in all baseline comparisons.

\begin{figure}[htbp]
    \centerline{\includegraphics[width=\linewidth]{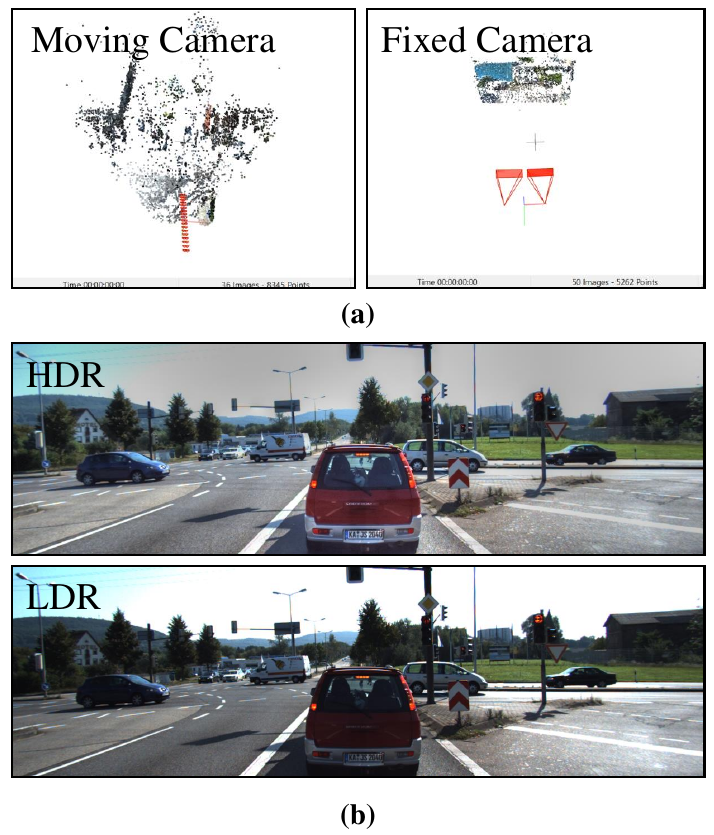}}
    \vspace{-10px}
    \caption{\textbf{(a)} Visualization of sparse point clouds initialized using COLMAP for fixed and moving cameras respectively, with the red triangular pyramid representing camera poses. \textbf{(b)} Comparison between Low Dynamic Range (LDR)  and High Dynamic Range (HDR) image processed by \textit{Photomatix}, showcasing the variation in scene illumination and tone mapping.}
    \label{fig:KITTI_Description}
    \vspace{-20px}
\end{figure}
\noindent\textbf{KITTI}~\cite{menze2015object} dataset contains 400 dynamic scenes from the KITTI raw data and we selected 5 scenes with 40 frames (20 from each camera) per scene for dataset construction. One viewpoint is used for training, while the other serves as validation.
The five scenes include two captured from fixed cameras (parked) and three from moving cameras (in transit), as shown in Fig.~\ref{fig:KITTI_Description} (a). Furthermore, we observed that in autonomous driving scenarios, low dynamic range (LDR) images captured due to direct sunlight significantly increase the difficulty of scene reconstruction. To mitigate this issue, we applied high dynamic range (HDR) processing to the images in one of the scenes using \textit{Photomatix}\footnote{Photomatix: \url{https://www.hdrsoft.com/}}, serving as a reference for reconstruction in autonomous driving scenarios that utilize high dynamic range images, as shown in Fig.~\ref{fig:KITTI_Description} (b). All scenes are selected at the maximum resolution corresponding to the Ground Truth, with sparse point clouds initialized using COLMAP.

\begin{table*}[htbp]
\caption{
\textbf{Quantitative results on the synthesis dataset D-NeRF~\cite{pumarola2021d}.} The \textbf{best} results are highlighted in bold. The rendering resolution is set to $800\times800$.
}
\begin{center}
\resizebox{0.95\textwidth}{!}{
\begin{tabular}{lcccccc}
\Xhline{2pt}
Method                                                 & \# Volume   & PSNR (dB)↑     & SSIM ↑        & LPIPS ↓       & FPS ↑         & Training Time (min) ↓ \\ \hline
K-Planes~\cite{fridovich2023k}   & CVPR’23     & 32.61          & 0.97          & -             & 0.97          & 52.0                  \\
V4D~\cite{gan2023v4d}            & TVCG’23     & 33.72          & 0.98          & 0.02          & 2.08          & 414.                  \\ \hline
4DGS~\cite{wu20244d}             & CVPR’24     & 34.14          & 0.94          & 0.02          & 84.4          & 40.2                  \\
RealTime4DGS~\cite{yang2023real} & ICLR’24     & 30.55          & 0.94          & 0.06          & 100.          & 37.1                  \\
Spacetime~\cite{li2024spacetime} & CVPR’24     & 29.76          & 0.95          & 0.04          & \textbf{113.} & \textbf{10.0}         \\
4DRotor~\cite{duan20244d}        & SIGGRAPH’24 & 31.73          & \textbf{0.97} & 0.03          & 47.2          & 102.                  \\
\rowcolor[HTML]{EFEFEF} 
SDD-4DGS (Ours)                                                   & -           & \textbf{34.82} & 0.96          & \textbf{0.03} & 85.3          & 46.1                  \\ \Xhline{2pt}
\end{tabular}
}
\end{center}
\label{table:d-nerf app}
\end{table*}



\begin{figure*}[htbp]
    \centerline{\includegraphics[width=\linewidth]{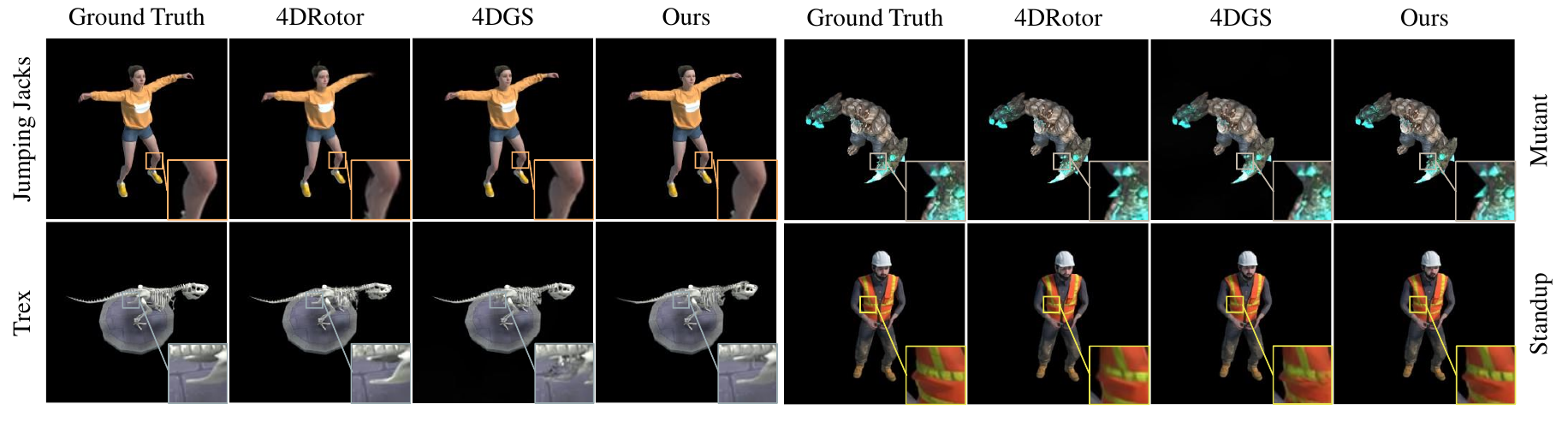}}
    \vspace{-15px}
    \caption{\textbf{Qualitative comparison between rendered results of D-NeRF~\cite{pumarola2021d} dataset.} We visualize the rendering results of our method with those of other methods~\cite{duan20244d, wu20244d} and enlarge the local details.}
    \label{fig:dnerf}
    \vspace{-15px}
\end{figure*}

\section{Additional Experiments}
In this section, we provide detailed performance metrics of subdivision rendering across different scenarios for further reference. Furthermore, we visualize the rendering results for both D-NeRF~\cite{pumarola2021d} and KITTI~\cite{menze2015object} scenarios separately.

\subsection{Monocular Synthetic Scene.}
We test the 8 scenes of D-NeRF~\cite{pumarola2021d} dataset. Due to the simple motion patterns and uniform lighting conditions in the synthesized dataset, there are insignificant differences among the different rendering techniques. Nonetheless, SDD-4DGS achieves a PSNR rendering quality of 34.82, exceeding the previous best of 34.41 as shown in Tab.~\ref{table:d-nerf app}. Additionally, it attained a rendering speed of 85.3 frames per second at a resolution of $800 \times 800$. The static-dynamic aware decoupling reconstruction strategy prevents occlusions in dynamic segments from causing blurring or loss of details in static components. This method facilitates better capture of intricate textures in stationary regions, as illustrated in Fig.~\ref{fig:dnerf}.
For areas exhibiting minimal or slight motion, SDD-4DGS guarantees high-quality reconstruction of dynamic parts while effectively characterizing fine features such as the textures of stationary clothing and the edges of rigid structures.

\textbf{Efficiency Experiment:}We evaluated the FPS and training time of various baseline algorithms on the D-nerf dataset. All experiments were conducted in a unified environment using a single NVIDIA RTX 3090 GPU for training and rendering. Since 4DRotor~\cite{duan20244d} did not release its CUDA implementation, we used its Torch version for comparison. The experimental results indicate that Spacetime~\cite{li2024spacetime} achieves the fastest training speed and the highest FPS due to its efficient polynomial fitting approach. Other algorithms exhibit negligible differences in training time and FPS. Additionally, for 4DGS~\cite{wu20244d}, the inclusion of dynamic-static awareness mechanisms results in a reasonable increase in training time.

\subsection{Multi-view Real-world Scene.}
We employ the Neu3D's~\cite{li2022neural} dataset as the benchmark for evaluating our method in real-world settings. The Neu3D's~\cite{li2022neural} dataset includes multi-view videos of six real-world scenes featuring flowing liquids and reflective materials. We selected five scenes from this dataset and used point clouds generated by COLMAP as the initialization conditions for these scenes.
\begin{table*}[htbp]
\caption{
\textbf{Quantitative results on the multi-view real dataset Neu3D’s~\cite{li2022neural}.} The \textbf{best} results are highlighted in bold. The rendering resolution is set to $1352\times1014$.
}
\begin{center}
\resizebox{0.98\textwidth}{!}{
\begin{tabular}{lccccccccc}
\Xhline{2pt}
                                                       & \multicolumn{3}{c}{\textbf{Coffee Martini}}               & \multicolumn{3}{c}{\textbf{Cook Spinach}}                 & \multicolumn{3}{c}{\textbf{Cut Roasted Beef}}             \\
\multirow{-2}{*}{\textbf{Neu3D’s}}                     & PSNR (dB) $\uparrow$ & SSIM $\uparrow$ & LPIPS $\downarrow$ & PSNR (dB) $\uparrow$ & SSIM $\uparrow$ & LPIPS $\downarrow$ & PSNR (dB) $\uparrow$ & SSIM $\uparrow$ & LPIPS $\downarrow$ \\ \hline
4DGS~\cite{wu20244d}             & 27.31                & 0.904           & 0.134            & 31.80                & 0.944           & 0.112            & 32.16                & 0.944           & 0.119            \\
RealTime4DGS~\cite{yang2023real} & 27.75                & 0.915           & 0.118            & 32.31                & 0.954           & 0.089            & \textbf{33.35}       & \textbf{0.957}  & \textbf{0.087}   \\
Spacetime~\cite{li2024spacetime} & 27.99                & 0.908           & 0.133            & 30.53                & 0.940           & 0.123            & 30.89                & 0.941           & 0.128            \\
4DRotor~\cite{duan20244d}        & 28.11                & 0.912           & 0.127            & 32.29                & 0.950           & 0.099            & 32.44                & 0.951           & 0.097            \\
\rowcolor[HTML]{EFEFEF} 
SDD-4DGS (Ours)                                        & \textbf{29.03}       & \textbf{0.917}  & \textbf{0.112}   & \textbf{32.83}       & \textbf{0.954}  & \textbf{0.089}   & 33.20                & 0.955           & 0.094            \\ \Xhline{1pt}
                                                       & \multicolumn{3}{c}{\textbf{Flame Steak}}                  & \multicolumn{3}{c}{\textbf{Sear Steak}}                   & \multicolumn{3}{c}{\textbf{Mean}}                         \\
\multirow{-2}{*}{Neu3D’s}                              & PSNR (dB) $\uparrow$ & SSIM $\uparrow$ & LPIPS $\downarrow$ & PSNR (dB) $\uparrow$ & SSIM $\uparrow$ & LPIPS $\downarrow$ & PSNR (dB) $\uparrow$ & SSIM $\uparrow$ & LPIPS $\downarrow$ \\ \hline
4DGS~\cite{wu20244d}             & 31.87                & 0.946           & 0.106            & 32.98                & 0.950           & 0.105            & 31.23                & 0.938           & 0.115            \\
RealTime4DGS~\cite{yang2023real} & 32.63                & 0.961           & 0.077            & 33.14                & \textbf{0.964}  & \textbf{0.076}   & 31.84                & 0.950           & \textbf{0.089}   \\
Spacetime~\cite{li2024spacetime} & 30.18                & 0.944           & 0.114            & 32.26                & 0.952           & 0.102            & 30.37                & 0.937           & 0.120            \\
4DRotor~\cite{duan20244d}        & 33.11                & 0.958           & 0.085            & 32.72                & 0.958           & 0.086            & 31.73                & 0.946           & 0.099            \\
\rowcolor[HTML]{EFEFEF} 
SDD-4DGS (Ours)                                        & \textbf{32.94}       & \textbf{0.961}  & \textbf{0.076}   & \textbf{33.98}       & 0.962           & 0.078            & \textbf{32.39}       & \textbf{0.950}  & 0.090            \\ \Xhline{2pt}
\end{tabular}
}
\end{center}
\label{table:n3v_detail}
\vspace{-10px}
\end{table*}

Our results, summarized in Tab.~\ref{table:n3v_detail}, demonstrate that our SDD-4DGS achieves superior rendering quality compared to other baseline approaches across diverse scenarios and on average. This performance is particularly pronounced in Scene \textit{Coffee Martini}, where daylight conditions introduce a complex interplay of light and shadow, presenting significant challenges for most methods. Daytime indoor scenes are heavily influenced by natural light, with variations in sunlight intensity, angle, and diffuse light distribution leading to intricate global illumination effects and localized highlights. These phenomena compromise color consistency and hinder the recovery of geometric details during reconstruction. Moreover, strong direct sunlight often causes specular reflections and glare on reflective surfaces such as glass, metal, and tiles, further distorting color information and inducing depth map noise and data loss. Despite these challenges, our method achieves substantial improvements over the previous state-of-the-art (SOTA), demonstrating its robustness in complex scenarios.


\subsection{Monocular Real-World Urban Scene.}
\begin{table*}[htbp]
\caption{
\textbf{Quantitative results on the multi-view real dataset KITTI~\cite{menze2015object}.}
Scenes are classified by High Dynamic Range (LDR) and Low Dynamic Range (LDR) imaging conditions with fixed cameras (FC) or moving cameras (MC).
The \textbf{best} results are highlighted in bold. The rendering resolution is set to $1242\times 375$.
}
\begin{center}
\resizebox{0.98\textwidth}{!}{
\begin{tabular}{@{}lccccccccc@{}}
\Xhline{2pt}& \multicolumn{3}{c}{\textbf{Scene 1 (LDR \& MC)}}                        & \multicolumn{3}{c}{\textbf{Scene 2 (LDR \& FC)}}                                                & \multicolumn{3}{c}{\textbf{Scene 3 (LDR \& FC)}}                                                       \\
\multirow{-2}{*}{\textbf{KITTI }} & PSNR (dB) $\uparrow$ & SSIM $\uparrow$ & LPIPS $\downarrow$ & PSNR (dB) $\uparrow$ & SSIM $\uparrow$ & LPIPS $\downarrow$ & PSNR (dB) $\uparrow$         & SSIM $\uparrow$              & LPIPS $\downarrow$           \\ \midrule
4DGS~\cite{wu20244d}                     & 26.27                & 0.894           & 0.099              & 29.10                & 0.939           & 0.054                                      & {\color[HTML]{000000} 16.09} & {\color[HTML]{000000} 0.421} & {\color[HTML]{000000} 0.450} \\
RealTime4DGS~\cite{yang2023real}         & 23.44                & 0.804           & 0.348              & 23.50                & 0.783           & 0.174                                      & 17.55                        & 0.672                        & 0.312                        \\
Spacetime~\cite{li2024spacetime}         & \textbf{27.12}       & 0.887           & 0.107              & 27.77                & 0.896           & 0.090                                      & 18.09                        & 0.752                        & 0.253                        \\
4DRotor~\cite{duan20244d}                & 19.16                & 0.675           & 0.277              & 10.20                & 0.344           & 0.876                                      & 8.69                         & 0.371                        & 0.648                        \\
\rowcolor[HTML]{EFEFEF} 
SDD-4DGS (Ours)                          & 26.77                & \textbf{0.901}  & \textbf{0.089}     & \textbf{29.80}       & \textbf{0.948}  & \textbf{0.044}                             & \textbf{23.54}               & \textbf{0.925}               & \textbf{0.063}               \\
\Xhline{1pt}  & \multicolumn{3}{c}{\textbf{Scene 4 (HDR \& MC)}}                        & \multicolumn{3}{c}{\textbf{Scene 5 (LDR \& MC)}}                                                & \multicolumn{3}{c}{\textbf{Mean}}                                                          \\
\multirow{-2}{*}{\textbf{KITTI }} & PSNR (dB) $\uparrow$ & SSIM $\uparrow$ & LPIPS $\downarrow$ & PSNR (dB) $\uparrow$ & SSIM $\uparrow$ & LPIPS $\downarrow$                         & PSNR (dB) $\uparrow$         & SSIM $\uparrow$              & LPIPS $\downarrow$           \\ \midrule
4DGS~\cite{wu20244d}                     & 14.65                & 0.549           & 0.694              & 20.15                & 0.769           & 0.179                                      & 21.25                        & 0.71                         & 0.30                         \\
RealTime4DGS~\cite{yang2023real}         & 17.63                & 0.651           & 0.311              & 19.07                & 0.702           & 0.447                                      & 20.24                        & 0.72                         & 0.32                         \\
Spacetime~\cite{li2024spacetime}         & 16.16                & 0.613           & 0.627              & 19.44                & 0.723           & 0.210                                      & 21.72                        & 0.77                         & 0.26                         \\
4DRotor~\cite{duan20244d}                & 16.63                & 0.593           & 0.746              & 15.11                & 0.506           & 0.303                                      & 13.96                        & 0.498                        & 0.570                        \\
\rowcolor[HTML]{EFEFEF} 
SDD-4DGS (Ours)                          & \textbf{21.11}       & \textbf{0.870}  & \textbf{0.119}     & \textbf{20.97}       & \textbf{0.788}  & \textbf{0.158}                             & \textbf{24.44}               & \textbf{0.89}                & \textbf{0.09}                \\
\Xhline{2pt}               
                                                 
\end{tabular}
}
\end{center}
\label{table:kitti_detail}
\vspace{-10px}
\end{table*}

\begin{figure*}[!htbp]
    \centerline{\includegraphics[width=\linewidth]{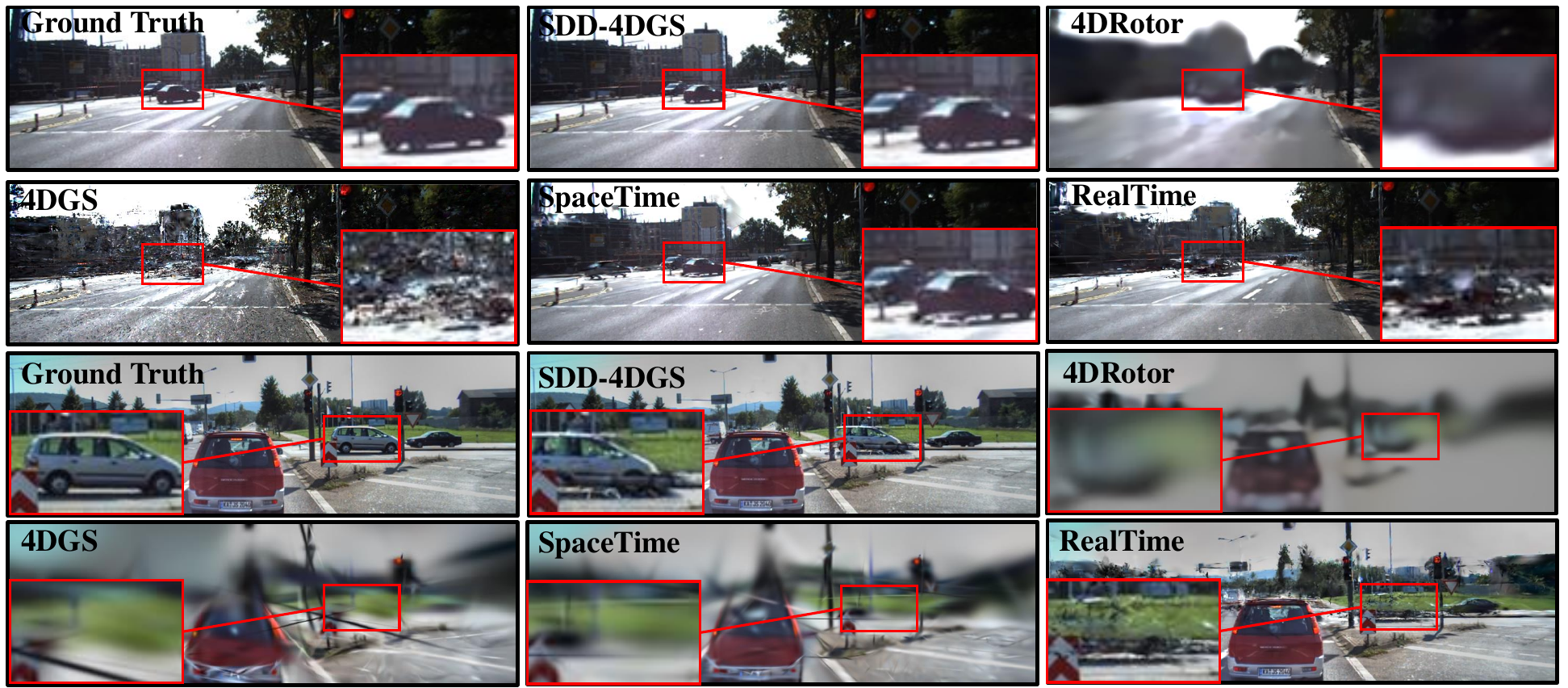}}
    \caption{\textbf{Comparative visualization of Gaussian splatting techniques on the KITTI~\cite{menze2015object} dataset.} The images highlight SDD-4DGS's enhanced detail preservation in dynamic objects like vehicles, closely matching the ground truth with sharp edges and accurate textures. In contrast, baseline methods such as 4DGS and SpaceTime show significant blurring and artifacts. These results demonstrate the robustness of SDD-4DGS against the challenges of intense outdoor lighting and motion blur, underscoring its potential for reliable vehicle detection and tracking.}
    \label{fig:KITTI}
\end{figure*}

We extended the KITTI~\cite{menze2015object} dataset to explore the applicability of Gaussian Splatting~\cite{kerbl20233d} technology in real urban environments, particularly within autonomous driving applications. Compared to indoor data, urban datasets typically face challenges such as more intense natural lighting, restricted camera viewpoints, significant variations in camera poses, and motion blur along with sparse observations due to fast-moving objects. SDD-4DGS demonstrates superior performance compared to other baseline methods in handling scenes with numerous dynamic components and static backgrounds, particularly in terms of overall image quality metrics such as PSNR and SSIM in Tab.~\ref{table:kitti_detail}. Additionally, some baseline methods may experience training failures under extreme conditions. SDD-4DGS demonstrated an exceptional ability to preserve details in dynamic objects, such as vehicles. As illustrated in Fig.~\ref{fig:KITTI}, SDD-4DGS maintained sharp edges and textural details closely matching the ground truth, essential for accurate vehicle detection and tracking. In contrast, methods like 4DGS and SpaceTime exhibited significant blurring and artifacts around moving cars, which could detrimentally impact the performance of perception systems in autonomous vehicles.

\subsection{Automatic Supervision Signal}
\label{subsec:uncertainty_supervision}

To facilitate effective static-dynamic decoupling in 4D reconstruction, we introduce an uncertainty-based automatic supervision signal~\cite{kulhanek2024wildgaussians} that generates binary masks to guide the optimization process. This approach builds upon the observation that dynamic regions typically exhibit higher uncertainty during reconstruction due to their time-varying nature.

\noindent\textbf{Motion Mask Generation}
The key challenge in deriving supervision signals for static-dynamic decoupling lies in automatically identifying which regions in each frame correspond to dynamic or static components without explicit annotations. We address this through an uncertainty estimation mechanism that leverages the semantic consistency properties of pre-trained visual features.
Specifically, we utilize DINOv2~\cite{oquab2023dinov2} to extract patch-level features from both rendered and ground truth images. These features exhibit remarkable robustness to illumination changes while remaining sensitive to geometric inconsistencies—a critical property for distinguishing between static structures and dynamic objects. For each image pair, we define the uncertainty score $\sigma$ through a linear mapping of feature similarity:
\begin{equation}
\mathcal{L}_{uncertainty} = \min (1, 2 - 2\cos(\tilde{D}, D)) \frac{1}{2\sigma^2} + \lambda_{prior} \log \sigma
\end{equation}
, where $\tilde{D}$ and $D$ represent the DINOv2 features extracted from patches of the rendered and training images, respectively, with $\cos(\tilde{D}, D)$ denoting the cosine similarity between these features vectors. The term $\lambda_{prior}\log\sigma$ serves as a regularization prior to prevent uncertainty collapse.

The uncertainty scores are optimized independently from the main reconstruction parameters to prevent gradient interference, ensuring that the densification algorithm in the Gaussian splatting pipeline maintains its effectiveness. After obtaining the optimized $\sigma$ values, we convert them into binary masks using:
\begin{equation}
m = \mathbbm{1}\left(\frac{1}{2\sigma^2} > 1\right)
\end{equation}
, where $\mathbbm{1}$ is the indicator function that evaluates to 1 when $\sigma^2 < \frac{1}{2}$. This approach effectively identifies dynamic regions (where $m=1$) without requiring explicit motion annotations.
Additional details can be found in~\cite{kulhanek2024wildgaussians}.

\end{document}